\documentclass[preprint,12pt,authoryear]{elsarticle}

\usepackage{amssymb}

\usepackage[ruled, vlined, linesnumbered]{algorithm2e}
\usepackage{algpseudocode}
\usepackage{graphicx}
\usepackage{float}
\usepackage{xcolor}
\usepackage{url}
\usepackage{amsmath}
\usepackage{caption}
\usepackage{subcaption}
\usepackage{booktabs}
\usepackage{framed}
\usepackage{booktabs}
\usepackage{tabularx}
\usepackage{multirow}
\usepackage{makecell} 
\usepackage{balance} 
\usepackage{soul}
\usepackage{booktabs}
\usepackage{hyperref}
\usepackage{caption}
\usepackage{subcaption}
\usepackage{todonotes}
\hypersetup{%
  colorlinks=false,
  linkbordercolor=white,
  pdfborderstyle={/S/U/W 1}
}
\usepackage{spverbatim}
\newcommand{\tuple}[1]{\ensuremath{\langle #1 \rangle}}

\newcommand{\pre}{\textit{pre}}
\newcommand{\tabitem}{~~\llap{\textbullet}~~}

\usepackage{xcolor}

\newcommand{\revision}[1]{{\textcolor{black}{#1}}}

\newcommand{\revisiontwo}[1]{{\textcolor{black}{#1}}}

\newtheorem{example}{Example}

\journal{Artificial Intelligence Journal}

\begin{document}

\begin{frontmatter}



\title{A Domain-Independent Agent Architecture for Adaptive Operation in Evolving Open Worlds}


\author[PARC]{Shiwali Mohan$^*$}
\author[PARC]{Wiktor Piotrowski$^*$}
\author[PARC,BGU]{Roni Stern}
\author[PARC]{Sachin Grover}
\author[PARC]{Sookyung Kim}
\author[PARC]{Jacob Le}
\author[PARC]{Yoni Sher}
\author[PARC]{Johan de Kleer}
\affiliation[PARC]{organization={Palo Alto Research Center}, country={USA}}
\affiliation[BGU]{organization={Ben Gurion University}, country={Israel}}

\begin{abstract}
\def\thefootnote{*}\footnotetext{These authors contributed equally to this work.}\def\thefootnote{\arabic{footnote}}
Model-based reasoning agents are ill-equipped to act in novel situations in which their model of the environment no longer sufficiently represents the world. We propose HYDRA, a framework for designing model-based agents operating in mixed discrete-continuous worlds that can autonomously detect when the environment has evolved from its canonical setup, understand how it has evolved, and adapt the agents' models to perform effectively. HYDRA is based upon PDDL+, a rich modeling language for planning in mixed, discrete-continuous environments. It augments the planning module with visual reasoning, task selection, and action execution modules for closed-loop interaction with complex environments. HYDRA implements a novel meta-reasoning process that enables the agent to monitor its own behavior from a variety of aspects. The process employs a diverse set of computational methods to maintain expectations about the agent's own behavior in an environment. Divergences from those expectations are useful in detecting when the environment has evolved and identifying opportunities to adapt the underlying models. HYDRA builds upon ideas from diagnosis and repair and uses a heuristics-guided search over model changes such that they become competent in novel conditions. The HYDRA framework has been used to implement \emph{novelty-aware} agents for three diverse domains - CartPole++ (a higher dimension variant of a classic control problem), Science Birds (an IJCAI competition problem\footnote{\href{http://aibirds.org/}{Angry Birds Competition}}), and PogoStick (a specific problem domain in Minecraft). We report empirical observations from these domains to demonstrate the efficacy of various components in the novelty meta-reasoning process. 
\end{abstract}




\end{frontmatter}


\section{Introduction}
\label{sec:introduction}

Artificial Intelligence (AI) and Machine Learning (ML) research on sequential decision-making usually rely on the assumption that key characteristics of the environment remain fixed after the decision-making agent is deployed.  
For model-based reasoning agents (e.g., based on Automated Planning), the fixed environmental characteristics are encoded explicitly in the domain model as actions, events, and processes that govern the agent's beliefs about the environment's dynamics. 
\revisiontwo{Model-based agents that are designed for stochastic and partially observable environments, such as POMDP, still assume the state transition and observation functions follow a fixed distribution.}
In model-free learning agents (e.g., based on Deep Reinforcement Learning), the fixed environmental characteristics are encoded implicitly in the simulations from which the action selection policy is learned. 
This assumption --- fixed environment characteristics --- poses a significant challenge in deploying intelligent agents. 
Model-based agents using domain knowledge that is outdated or incorrect can cause the agent to fail catastrophically during deployment. 
Model-free learning agents will need numerous interactions with the environment to learn a new policy, rendering them ineffective when the environment evolves from what they were trained on. 
\revision{Other challenges posed by the real world (e.g., partial observability and sensing noise, non-deterministic actions,  continuous spatio-temporal dynamics, etc.) have received significant attention in the literature over the years, while the major challenge of robust operation in an open and evolving world has been largely overlooked until recently.} 

This paper studies how to design intelligent agents that can robustly operate in an \emph{open and evolving world}, an environment whose characteristics change unexpectedly while the agent is operational. Such a shift in environmental characteristics has been referred to as a \emph{novelty}~\citep{boult2021towards,langley2020open}. An effective \emph{open world} agent can autonomously \emph{detect} when a novelty has been introduced in the environment, \emph{characterize} it, as it pertains to what it knows about the environment, and then \emph{accommodate} it by changing its decision-making strategies. Ideally, it \emph{transfers} relevant operational knowledge from before novelty is introduced to after, i.e., it learns without fully retraining and in orders of magnitude less time. This challenge of designing such an open-world agent, where novelties can appear at unspecified time, has been gaining significant interest in the AI literature \revisiontwo{\citep{senator2019sailon, AI_for_openworlds_2022,langley2020open,liu2023ai,xu2024active,goel2021novelgridworlds,boult2022weibull}.}

This paper advances the \emph{AI for open world} research agenda by studying how model-based reasoning agents can reason about novelties appearing in open worlds and adapt themselves in response. A key characteristic of model-based reasoning agents is that their models of the environment are \emph{explicit} and \emph{compositional}. Each element of the model represents a meaningful aspect of environmental dynamics. Consider a planning agent written in
PDDL+~\citep{fox2006modelling} to aim a ball at a target. It will include formal specifications of processes such as the movement of a ball under the effect of gravity. This process is encoded separately from other aspects of the environment, such as the ball bouncing off a hard surface. Together, all elements in the PDDL+ model determine the agent's beliefs about environment dynamics. When considering open-world agent design, compositional models have significant advantages when compared to black-box, end-to-end, integrative models such as deep neural networks. They generate explicit expectations about future outcomes that are expressed in meaningful terms (e.g., gravity). This enables a focused analysis of what might have changed in the environment. Often, the introduced novelty only impacts a small subset of model elements and only those need to be updated. Consequently, model-based learning agents can adapt to novelties with fewer observations than model-free agents. 

The paper introduces HYDRA, \revision{a framework for designing model-based autonomous agents that can operate in complex open worlds. At their core, HYDRA agents use explicit planning models of their environment for action reasoning and selection. The planning models are described in PDDL+, an expressive and feature-rich modeling language. PDDL+ supports reasoning with mixed discrete-continuous state and action spaces. Further, it enables nuanced reasoning about exogenous environment behavior, i.e., events and processes that are independent of the agent's control. Building upon PDDL+ enables us to study complex environments where transition dynamics are governed by both the agent and the environment.}

HYDRA augments PDDL+ planning with visual reasoning, task selection, and action execution to develop agents that operate in a closed loop with the environment. A key contribution of our work is a meta-reasoning process for novelty that is integrated with the basic agent \emph{perceive-decide-act} loop. The meta-reasoning process monitors relevant aspects of agent behavior in the environment; including analyzing the observation space, tracking state changes in the environment, and monitoring the quality of performance. When the environment evolves from the canonical setup the agent is designed for, these monitors generate signals that trigger an adaptation cycle. \revision{PDDL+ model adaptation employs a heuristic search to identify which element(s) of the model needs to be revised. HYDRA's repair module makes explicit and relevant modifications to the elements identified, enabling the agent to improve its decision-making. During adaptation, HYDRA can update its PDDL+ model as well as change the order in which it executes tasks. }

HYDRA is domain-independent and has been used to design agents for three research domains - CartPole++ (a higher dimension variant of a classic control problem CartPole), ScienceBirds (an AI competition domain), and PogoStick (a Minecraft domain). Our results show that for certain types of novelties, HYDRA agents can adapt \emph{quickly} with few interactions with the environment. Additionally, the adaptations produced by HYDRA are \emph{interpretable} by design - they are represented in terms of changes to the elements of its model, enabling inspection of proposed changes. This property of a HYDRA agent is a considerable advantage when developing adaptive systems that can be trusted.


\section{Related Work}
\label{sec:related_work}



The problem we consider in this work is raised when a model-based reasoning agent fails to act in the environment because its underlying models are deficient or incorrect. The approach we propose to address this problem is to adapt the world model our agent is using based on the observations it collects. We begin by providing an introduction to PDDL+. Then, we summarize the related prior work on 
(1) planning under uncertainty, 
(2) handling execution failures in planning-based agents, 
(3) repairing world models for agents, 
(4) automated diagnosis and repair, 
and (5) learning planning models from observations. Then, we review the current state-of-the-art on open-world learning and identify how our research advances it.

\subsection{Domain Modeling and Planning in Complex Environments}
\revision{Autonomous agents often act in and reason with multifaceted environments governed by complex system dynamics. HYDRA uses PDDL+~\citep{fox2006modelling} to accurately and efficiently capture the relevant characteristics and behavior of said environments as planning domains. }

\revision{PDDL+ is a feature-rich language designed for defining mixed discrete-continuous systems as planning models. It significantly expands on the expressive power of previous versions of the planning modeling language, e.g., PDDL2.1 \citep{fox2003pddl2} and PDDL \citep{mcdermott1998pddl}. PDDL+ is capable of representing diverse system elements and dynamics via the use of such concepts as propositional and numeric state variables, instantaneous and durative actions, concurrent behavior, continuous action effects, or timed-initial activity. However, the most significant advancement in expressiveness is by enabling the modeling of an \emph{independent} environment that evolves dynamically over time. In every prior version of the PDDL language, the world only changed via the agent's actions. PDDL+ introduced constructs to model exogenous activity: discrete events (mode switches) and durative processes (continuous flows). In PDDL+ models, the agent can comprehensively reason about the evolution of its world but can only interact with it through defined actions. The agent cannot dictate whether events or processes are executed, instead, their effects are applied as soon as their preconditions are satisfied (i.e., 'must-happen' behavior). As an example from the Angry Birds domain, a collision between a bird and a wooden block is modeled by a PDDL+ event, while a process models the continuous ballistic flight of a bird launched from a slingshot. Such phenomena, especially physics-based, are ubiquitous in real-world applications. Thus PDDL+ facilitates the capturing of realistic scenarios as planning domains much more accurately, compared to other standardized planning modeling languages.}

\revision{PDDL+ planning is a notoriously challenging task that belongs to the undecidable class of problems. PDDL+ models of real-world scenarios contain a wide range of features, vast state spaces, and complex (usually non-linear and/or discontinuous) system dynamics. Therefore, modeling accuracy must be balanced with efficiency to ensure that the model is adequately detailed for real-world applications and that the solution is generated in a reasonable time. Furthermore, to effectively reason with all classes of novelty, the PDDL+ models must be general enough to accurately capture complex system components and behaviors, but also flexible enough to accept nuanced model updates to effectively adapt to novelty. However, the space of possible domain shifts is infinite. The PDDL+ model should be designed to explicitly model as many relevant aspects of the environment as is feasible to account for environment components that can be affected by the introduced novelties. However, this computational overhead should be minimized to ensure the efficiency and solvability of the resulting PDDL+ model.}

\revision{PDDL+ domains are fully observable and must explicitly encode all of their components and variables. Consequently, they are best suited for representing fully observable and deterministic dynamical systems. Simulation environments, on the other hand, usually only reveal part of their system variables as perception/sensor data. Thus, the PDDL+ model is partially built on assumptions made by the designer about some of the underlying structure and dynamics of the system. Similarly, fully-observable PDDL+ domains can model partially-observable environments by incorporating explicit assumptions about the obscured parts of the system. However, the main advantage of using symbolic models is that they can be thoroughly investigated and explicitly modified to incorporate novelty. Sub-symbolic approaches are opaque and uninterpretable, they cannot be easily modified to account for novelty without expensive, and often infeasible, retraining.}

\revision{There are multiple PDDL+ planners available \citep{cashmore2016compilation,coles2014pddl+,coles2012colin}. However, most of them are limited to models with linear dynamics only, do not support the whole set of PDDL+ features, or do not scale well. UPMurphi \citep{della2009upmurphi}, DiNo~\citep{piotrowski2016heuristic}, and ENHSP \citep{scala2016interval} can handle complex domains and were initially explored, but were not able to efficiently scale to our development domains.}

\revision{HYDRA employs Nyx~\citep{piotrowski2024realworld}, a customizable domain-independent PDDL+ planner that facilitates solving complex feature-rich planning problems via tailored approaches (e.g., custom heuristics or domain language extensions).}

\revisiontwo{\subsection{Sequential Decision-Making Under Uncertainty}}

\revisiontwo{Sequential decision-making and planning under various forms of uncertainty is a core topic in AI research. 
The Markov Decision Problem (MDP)~\citep{puterman1990markov} is one of the most widely used models for capturing planning problems in which the effects of actions can be stochastic. 
Partially Observable MDP (POMDP)~\citep{kaelbling1995partially}, is a well-known extension of MDP that also supports uncertainty over the current state. 
Many algorithms have been proposed to solve MDP and POMDP problems, based on Dynamic Programming~\citep{shani2013survey,sanner2010symbolic}, Heuristics Search~\citep{horak2018goal,kim2019pomhdp,silver2010monte}, and Machine Learning~\citep{bhattacharya2020reinforcement,wang2021deep}.}

\revisiontwo{Unfortunately, one cannot use these algorithms directly for our problem, because both models --- MDP and POMDP --- assume that the world is not evolving. 
In an MDP, the non-evolving world assumption is embodied by assuming that for every action we know the set of possible effects it may have and their probability (this is expressed in the MDP transition function). 
In a POMDP, the non-evolving world assumption is embodied also by the observation function, which specifies for every possible observation the probability it will be observed given the actual state of the agent. 
In our problem, we do not know a priori how and when the world will evolve, and which novelties might occur. Thus, different algorithms and models are required than off-the-shelf MDP and POMDP algorithms.}

\revisiontwo{Reinforcement Learning (RL) algorithms~\citep{sutton2018reinforcement} are designed for solving sequential decision-making problems in settings where we do not know the exact transition and observation probabilities. Popular RL algorithms such PPO~\citep{schulman2017proximal} and DQN~\citep{mnih2013playing} learn how to act, i.e., output a \emph{policy},  by interacting with the environment. A main assumption made by standard RL algorithms is that the environment dynamics do not change. Thus, after enough interactions with the environment the RL agent's policy remains mostly the same. This is often embodied by a 
\emph{learning factor} that determines how much the incumbent policy should change based on recent observations. The learning factor decreases as more information is gathered. 
In the novelty setting we consider, the occurred novelty may change the environment dynamics itself, which may cause previously trained policies ineffective. We demonstrate this empirically in Section~\ref{sec:evaluation}. Indeed, applying RL algorithms in an evolving environment, sometimes referred to as \emph{Lifelong Reinforcement Learning}, is known to be significantly challenging, ~\citep{isele2018selective,abel2018policy}.}

\subsection{Adapting to Execution Failure}
Planning models are usually deemed correct by design. Execution failures and other discrepancies observed during the execution of the generated plans are often attributed to partial observability or non-determinism of the target environment. 
\emph{Replanning} and \emph{Plan Repair} are the common approaches to such failures. 
Replanning~\citep{cushing2005replanning, bezrucav2022towards,nebel1995plan} methods attempt to generate a new solution to the problem, either from the very beginning or from the point of failure of the plan, by using updated information from the environment. 
Plan repair~\citep{myers1999cpef, bidot2008plan, komenda2014domain,fox2006plan} methods adapt the plan according to additional data so that it will then be able to achieve the desired goal.
\revision{Replanning and plan repair algorithms usually assume that information about the environment that caused the plan execution failure is freely available and can be queried at any time. Put differently, replanning and plan repair assume that the agent's model is sufficiently correct and accurate but that the queried information (i.e., perception or sensor data) was noisy or altered by a rare phenomenon. Thus, replanning with updated perception data or fixing a single plan instance is sufficient to restore adequate performance in the presence of uncertainty.}


\subsection{Repairing World Models for Planning-Based Agents}

The Action Description Update (ADU) problem~\citep{eiter2010updating} is the problem of updating an action model given a set of new statements about the world dynamics, e.g., adding new axioms or constraints. This is different from learning how to update a model from observations. 
\citet{molineaux2012discoverhistory} used abductive reasoning about unexpected events to expand the knowledge base about the hidden part of the environment and improve their replanning process. This is more similar to handling partial observability than to repairing planning models to handle novelties in the environment. 

In the Model Reconciliation Problem~\citep{chakraborti2017plan,vasileiou2021exploiting,chakraborti2019plan} (MRP), plans generated by one agent (the planner) 
and the objective is to explain that plan to another agent (the observer). 
MRP has been studied in settings where each agent assumes a different world model, 
and the desired explanations are changes to the world model assumed by the observer. 
Generating such an explanation can be viewed as a form of model repair. 
However, they require knowledge of both models, while in our case we do not assume access to the world model after novelty has been introduced. 
To the best of our knowledge, all prior work on the ADU problem and MPR have dealt only with discrete domains while we proposed a general method for mixed discrete-continuous environments.

In the Model Maintenance Problem (MMP)~\citep{bryce2016maintaining}, 
a world model changes (drifts) over time, and the task is to adapt the model based on observations and active queries to minimize the difference between it and the real world. 
MMP has been studied in the context of planning agents, where the drifted model is a symbolic planning model. 
\citet{bryce2016maintaining} allowed queries about aspects of the planning model, e.g., ``is fluent $f$ a precondition of action $a$''. 
\citet{nayyar2022differential} allowed queries about the possible execution of plans, e.g., ``Can you perform the plan $a_1, a_2, a_3$'', where responses state which prefix of the given plan could be executed and the last state reached. 
Our problem can be viewed as a special case of MMP with a single drift (i.e., a single novelty) and our approach may be applicable to solve MMP. 
However, unlike prior work on MMP, we go beyond purely symbolic planning and support mixed discrete-continuous domains. Additionally, the only information available for model update are observations and the system cannot explicitly query for the structure of the true model.

~\citet{frank2015reflecting} discussed the challenge of adapting a planning model to novelties  
but did not propose a concrete approach to do so. 
~\citet{zhuo2013refining} proposed an algorithm for repairing a planning domain from observations, using a MAX-SAT solver.
~\cite{gragera2022repair} used a different approach to repair planning models where the effects of some actions are incomplete. Their approach compiles, for each unsolvable task, a new extended task where actions are allowed to insert the missing effects.  However, both works are limited to classical planning and cannot handle mixed discrete-continuous domains. 
Recent work has looked at how novelty can be accommodated in various discrete environments such as Polycraft \citep{muhammad2021novelty}, which is a mostly deterministic domain, and Monopoly \citep{loyall2022integrated,thai2022architecture}, which is a stochastic, strategic domain. 
Our research extends these lines of research and demonstrates that model revision techniques can support novelty accommodation in dynamic, physics-based domains such as CartPole and Science Birds.

Some prior work explored how to repair obsolete Markov Decision Problem (MDP) models based on observations~\citep{chen2013model,yang2022meaningful}. They proposed an approach that uses a model checker to ensure that the repaired MDP satisfies some necessary constraints. Along similar lines, \citet{pathak2015greedy} study the problem of repairing a discrete-time Markov Chain. Both approaches are not directly applicable to repairing rich planning models.

\subsection{Automated Diagnosis and Repair}
Automated diagnosis (DX) and repair of faulty systems is a core AI problem that deals with finding the root cause of the observed behavior and suggesting diagnostic and repair actions 
to return the system to a nominal state~\citep{dekleer2003fundamentals}. 
Many approaches and systems have been proposed in the past for DX and repair in both discrete~\citep{dekleer1987diagnosing} and mixed continuous-discrete settings~\citep{niggemann10model}. 
For example, \citet{sun1993aFramework} and ~\citet{barriga2020extensible} proposed general frameworks for repairing faulty systems by planning repair and diagnostic actions.
\citet{dekleer1987diagnosing} proposed the well-known General Diagnosis Engine which has been extended and improved by many~\citep{matei2018modelBased,feldman2020efficient} including to hybrid systems~\citep{niggemann2015diagnosis}, 
and software~\citep{abreu2011simultaneousDebugging,elmishali2018AnArificial}. 
While it might be possible to reduce our problem to a DX problem, where the system to repair is the agent's model, 
it is not clear whether existing DX and repair methods would work, and such reduction is not trivial.

\subsection{Learning Planning Models from Observations}
There is a growing literature on learning planning models from observations~\citep{arora2018review}, including algorithms such as ARMS~\citep{yang2007learning}, LOCM~\citep{cresswell2013acquiring}, LOCM2~\citep{cresswell2011generalised}, AMAN~\citep{zhuo2013action}, FAMA~\citep{aineto19}, and SAM~\citep{stern2017efficientAndSafe,juba2021safe,juba2022learning}. 
\citet{aineto2022comprehensive} arranged these algorithms in a comprehensive framework. 
However, none of these algorithms is designed to learn complex mixed discrete-continuous planning models such as the domains we consider in this work.  
Additionally, these algorithms are designed to learn a new planning model from scratch at training time, as opposed to repairing an existing planning model to novelties at run time.

~\citet{niggemann2012learning} proposed an algorithm for learning systems that can be captured as hybrid timed automata. 
While a PDDL+ domain and problem can be compiled into a hybrid timed automata, it is not clear how to transfer their approach to the problem of learning a complete PDDL+ domain, 
and whether such an approach can scale. 

\subsection{Open World Learning}
There is a growing body of work on AI for open worlds and novelty reasoning. \citet{senator2019sailon} introduced the open-world learning problem highlighting why it is a critical need of intelligent systems operating in real worlds and theorized the construct of \emph{novelty} --- a well-known concept in the human world --- for AI agents. \citet{langley2020open} elaborated on what kinds of novelties can exist in an open world, and \citet{doctor2022toward} study how domain complexity can be used to measure the difficulty of adapting to various novelties. \citet{boult2021towards} formally define the construct of novelty for perception problems. Our work contributes to the growing scientific understanding of open-world learning by proposing how novelties are defined for sequential decision-making agents. 

Open-world learning approaches have been studied for two different problems: perception \citep{dietterich2022,pang2022} and sequential decision-making --- our area of study. \citet{muhammad2021novelty} introduce a novelty handling framework for planning agents that can reason about and adapt to a class of novelties in discrete and deterministic environments. The novelties they study are those that can be observed as a direct change in the agent's observation space. \citet{goel2022rapid} extends this work for reinforcement learning agents, albeit still operating in discrete deterministic worlds and focusing on directly observable novelties. \citet{loyall2022integrated} study novelty reasoning methods for discrete, probabilistic environments. Our work advances this line of research in two dimensions; one, the agent framework introduced here operates in mixed discrete-continuous environments and second, the adaptation methods included in the framework can handle an extended class of novelties including those that don't impact the observation space directly but manifest as a change in environment transitions. \citet{musliner2021openmind} study novelty reasoning with discrete planning models for discrete and continuous environments. They propose a set of hypothesis-driven strategies that enable robust operations in novel conditions. Our framework is different in that we also adapt the planning models to reflect what has changed in novel conditions. Our own work \citep{piotrowski2023heuristic} proposes a model repair algorithm for CartPole-2D. This paper describes the general framework of which model repair is a part and presents results from multiple domains.

\section{Problem Setup and Research Domains}
\label{sec:problem}

\revision{We define \emph{novelty} as an unexpected and unknown domain shift that permanently alters the environment's composition, dynamics, or its interactions with the agent.}
\medskip

\revision{Formally, let an environment $E$ be a tuple $E = (S^*, A^*, H^*, F^*, G^*, O)$ where:
\begin{itemize}
    \item $S^*$ is a set of states that the environment can be in.
    \item $A^*$ is the set of agent's actions that can be executed in the environment.
    \item $H^*$ is the set of environmental happenings (analogous to actions) that encode the environment's dynamics beyond the agent's control.
    \item $F^*: S^* \times A^*\cup H^* \rightarrow S$ is the transition function (i.e., $F^*(s^*,a^*)\rightarrow s^{*\prime}$).
    \item $G^* \subseteq S^*$ is a set of goal states.
    \item $O$ is a function that generates an observation for a given state.
\end{itemize}
}

\revision{
A state of the environment $s^* \in S^*$ is a complete assignment of values over all its state variables at a time $t(s^*) \in \mathbb{R}$. The agent perceives the state of the environment via observation $O(s^*_{obs})$ where $s^*_{obs} \subseteq s^*$ is the observable part of the state of the environment. The agent then decides on an action $a^*\in A^*$ which modifies the state of the environment. The environment can evolve on its own via environmental happenings $h^* \in H^*$, which are analogous to the agent's actions. However, the agent cannot directly control the environment's dynamics but must reason with it before acting. $G^* \subseteq S^*$ is the set of goal states (i.e., a set of states that satisfy the goal conditions) that the agent is trying to reach\footnote{Note that $G^*$ is a specific formulation to designate desirable states and is used here because the paper studies model-based reasoning agents. A reward function $R^*$ may be used instead for reinforcement learning agents.}.}


\medskip

\revision{A \textbf{novelty} is a function $\nu(E) \rightarrow E'$ that takes the environment $E$ as input, modifies one or more of its components $(S^*, A^*, H^*, F^*, G^*, O)$, and returns a novel environment $E'$.}

\revision{Novelty can impact the agent-environment relationship in different ways. Some novelties can change the structure of the environment (i.e., modify $S^*$ or $O$) by introducing or removing novel object types, introducing new attributes of existing objects, or shifting the distribution of attribute values. Other novelties modify, introduce, or remove actions available to the agent or the environment, impacting $A^*$ or $H^*$, respectively. Novelties can also change the dynamics of state evolution by changing the transition function $F^*$. Finally, the goal conditions can be altered by the novelty, affecting $G^*$. Often, a single novelty changes multiple elements of the environment-agent relationship.} 



\revision{Open-world learning can be defined as an agent's ability to adapt to unexpected domain shifts in a dynamically evolving environment. To reason with the environment, model-based agents use an internal domain $D$ which is an approximation of environment $E$, i.e., $D \approx E$. After introducing novelty $\nu$, the agent's model $D$ becomes inconsistent with the updated environment $E'$, such that $D \not\approx E'$. \emph{Novelty detection} is the process of recognizing that a domain shift has occurred, i.e., the environment is now in $E'$. \emph{Novelty Characterization} is the process of determining the difference between the nominal environment $E$ and novel environment $E'$ in terms of elements of the model $D$. \emph{Novelty accommodation} is a process of adapting the agent's internal model $D$ to be consistent with the novelty-affected environment $E'$, such that the updated model $D' \approx E'$.}

\revision{In an open-world setting, novelty is considered from the agent's perspective, i.e., it impacts the relationship between the agent and the environment. In other words, novelty is a domain shift that the agent is unprepared for because: the novelty is not part of the agent's default internal model of the world or its internal model is significantly different from the true environment post-novelty; the agent has not previously experienced the novelty phenomenon, nor was it part of the agent's training. In all cases of meaningful novelty, further learning or model adaptation must occur for the agent to successfully perform in the novel world.}

\revision{It is important to make a distinction between novelty, i.e., a persistent domain shift, and uncertainty about the environment or outlier phenomena. Uncertainty is exhibited by an environment by default, even in a non-novel setting, and must be accounted for in the agent's internal model. 
Put differently, uncertainty is a characteristic of the environment, whereas novelty is an \emph{explicit change} in the environment characteristics. For instance, the transition function $F^*$, when defined in probabilistic terms, encodes the uncertainty in state transitions. On the other hand, novelty modifies the environment characteristics, thus, if the probabilities change in the transition function $F^*$, it would be considered a novelty in our framing. Partial observability is also distinct from novelty. As described above, the agent defines its internal model on partial observations $O(s_{obs}\subseteq s^*)$ about the state of the environment $E$. Novelty is a different concept entirely which can alter the observation function $O$ by altering its definition or adding noise, and so on.
}

In an open world, any number of novelties can present themselves at any time-point. In this paper, we limit our scope by making the following assumptions: (1) only specific classes of novelties are studied; (2) at most one novelty is introduced in a trial; (3) the novelty under study is introduced only between episodes; (4) once the novelty is introduced, it persists for the rest of the trial. We refer to this setup as the \emph{single persistent novelty setup}. These assumptions were made to support the iterative development of agents and allow clear experimental measurement\footnote{\revision{The single persistent novelty assumption is motivated by and grounded in automated diagnosis where developing faults are analogous to unexpected domain shifts. From a simple analysis, if components fail independently, a double fault is far less likely to occur than a single fault. Said differently, the sum of the probabilities of all single faults is far larger than that of all double faults. Double faults are very rare, if components fail independently, and become prohibitively complex to identify and disambiguate, especially in mixed discrete-continuous environments such as cyber-physical systems.}}.

In our setup, an agent interacts with the environment repeatedly for independent $N$ episodes that constitute a \emph{trial}. Each episode begins in some initial state of the environment. 
During the episode, the agent iteratively executes actions and observes their outcomes until a terminal state is reached, upon which the episode ends and a new one begins. After an unknown number $k$ of episodes, a novelty is introduced in the environment, requiring the agent to adapt. The agent is not informed about when the novelty is introduced or how it has changed the environment. The novelty persists for the rest of the trial, i.e., for the remaining $N-k$ episodes. 

\subsection{Research Domains}
Implementation and experiments described in this paper are motivated by the following three domains. Each domain has been specifically created and maintained for evaluating open-world learning by external independent research groups.

\subsubsection{CartPole++} 
\begin{figure}[t]
    \centering
    \includegraphics[width=0.85\columnwidth]{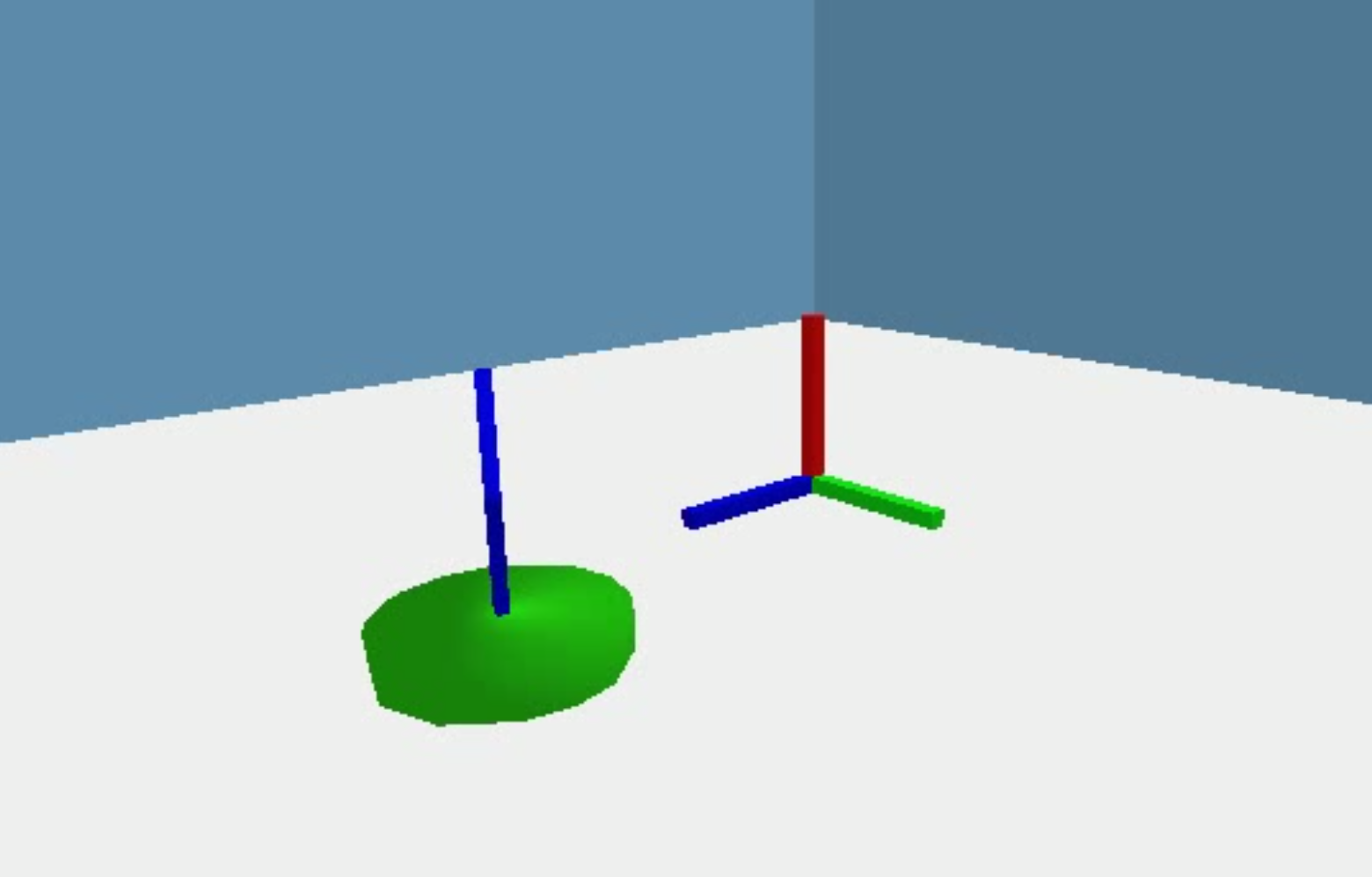}
    \caption{Screenshot of CartPole++ environment.}
    \label{fig:cartpole}
\end{figure}

This physics-based discrete-continuous domain 
is a higher dimensional version of the standard Reinforcement Learning benchmark problem CartPole. The agent can push the cart in any of the cardinal directions and the objective is to keep the pole upright for $200$ steps. 
Figure \ref{fig:cartpole} illustrates this domain. 
A variation of the domain additionally includes a set of spheres flying through the space.

\subsubsection{ScienceBirds} 
\begin{figure}
    \centering
    \includegraphics[width=0.6\columnwidth]{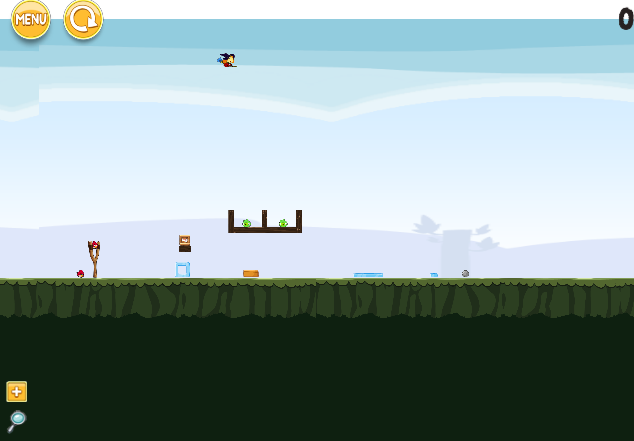}
    \caption{Screenshot of a ScienceBirds level}
    \label{fig:ab_level}
\end{figure}

This domain is a version of the popular video game Angry Birds, which has garnered widespread recognition over many years. The objective in SB involves eliminating all green pigs in a level while maximizing destruction to the surrounding structures. The player is equipped with a collection of birds to be launched from a slingshot, where the birds can vary in their special abilities (e.g., increased damage against specific block types, explode upon collision, etc.). The pigs are typically concealed within complex platform structures built with a variety of blocks, requiring the player to identify and eliminate the weak points of the structures, such as supports or dynamite. 
Figure \ref{fig:ab_level} shows a screenshot from this domain.  
The development of Science Birds utilized the Box2D open-source physics library, ensuring that all objects within the game's environment comply with the principles of Newtonian physics in a two-dimensional plane.

\subsubsection{PogoStick} 
\begin{figure}
\centering
\includegraphics[width=1\textwidth]{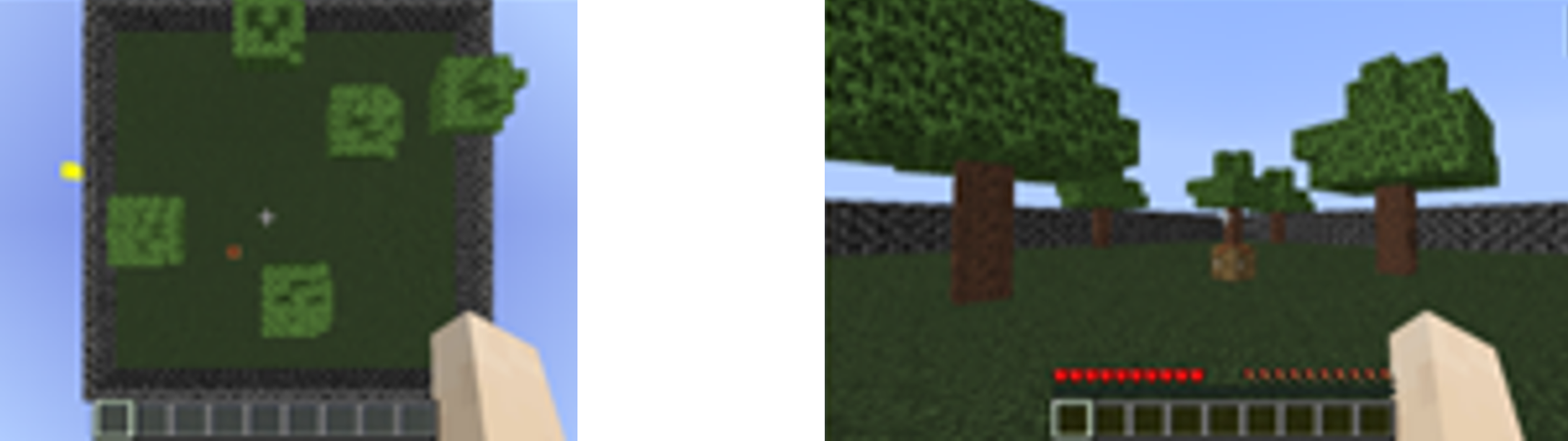}
\caption{Bird's-eye and a first-person view of the PogoStick environment.} 
\label{fig:polycraft_environment}
\end{figure}

This domain is a part of the Polycraft domain (developed by the University of Texas, Dallas\footnote{\url{https://github.com/StephenGss/PAL/}}), that is based on a version of the popular video game Minecraft\footnote{\url{https://www.minecraft.net/}}. Figure \ref{fig:polycraft_environment} shows the top- and first-person view of the PogoStick environment. In this domain, the task is to craft a pogo-stick given a recipe, that requires both short- and long-term planning and decision-making. The agent controls a character (Steve) who needs to navigate the environment, gather different resources the recipe requires, construct intermediate objects identified in the recipe, and interact with other agents present in the environment to trade. The agent succeeds if it crafts a PogoStick within the allotted time. PogoStick is a complex environment that is partially observable (unknown adjacent rooms that may hold relevant items) and multiagent (interacting with traders and competing for resources with rival pogoists).



\bigskip

Our research domains are mixed discrete-continuous and require complex goal-oriented, spatio-temporal reasoning and learning. CartPole++ and ScienceBirds are fully observable, deterministic, and physics-based - the environment dynamics are governed by physical laws such as flight under gravity. PogoStick, on the other hand, is discrete but partially observable and has some non-determinism. Additionally, PogoStick has other agent entities that have supportive or competitive intentions. The impact of an agent's action is immediately observable in CartPole++ and PogoStick. However, in ScienceBirds, actions have delayed consequences. The trajectory of the bird is determined very early in the process by setting angle and velocity, however, the consequences emerge much later after hitting structures. The differing characteristics of these domains pose a significant challenge in designing a common novelty-aware agent framework. Continuous space, physics-based dynamics of CartPole++ and ScienceBirds motivate accurate physics modeling in the agent framework. Partial observability in PogoStick motivates balancing information-gathering and execution needs. All domains require continual task monitoring and replanning for effective performance. Detecting and accommodating novelties is further complicated due to complex, temporal interactions between objects and entities in the domains.



\subsection{Space of Novelties}
Table \ref{tab:novelties} summarizes a subset of novelties that can be injected into the environments and examples of their instantiations in our research domains. An ideal novelty-aware agent can detect, characterize, and accommodate a wide range of novelties that impact the structure, dynamics, and constraints in the environment. Novelty IDs $1$ and $2$ impact the structure of the environment and may change in the observation space. Novelty IDs $2$, $3$, and $4$ impact specific transitions in the environment. $6$, $7$, and $8$ impact environmental transitions in general by changing the laws. Some novelty IDs ($2$, $7$) have been not instantiated in our research domains and consequently, are not being investigated currently. Our aim is to design a common novelty-aware agent framework that can reason about the full space of novelty. The sections below introduce our approach and summarize the progress we have made towards this goal. 

\begin{table}[ht]
\scriptsize
\begin{tabular}{@{}llp{2.5cm}p{2.5cm}p{2.5cm}p{2.5cm}@{}}
\toprule
ID &
  Type &
  Description &
  \multicolumn{3}{c}{Domain instantiation} \\
 &
   &
   &
  \multicolumn{1}{c}{CartPole++} &
  \multicolumn{1}{c}{ScienceBirds} &
  \multicolumn{1}{c}{PogoStick} \\ \midrule
1 &
  Attribute &
  New attribute of a known object or entity &
  Increased mass of cart &
  Increased bird launch velocity
  & Increased production of wooden logs
   \\
2 &
  Class &
  New type of object or entity &
  * &
  Orange bird &
  Thief steals resources from Steve
   \\
3 &
  Action &
  New type of agent behavior/control &
  * &
  * &
  * \\
4 & Interaction  & 
  New relevant interactions of agent, objects, entities & 
  Change in energy of object bounces & 
  Red birds can go through platforms &  
  Trees grown using saplings leave a new sapling when cut.\\
5 &
  Activity &
  Objects and entities operate under new dynamics/rules &
  Pole is attracted to blocks &
  Platforms move up and down &
  Traders provide the pogo-stick.
   \\
6 &
  Constraints &
  Global changes that impact all entities &
  Change in gravity &
  Change in gravity &
  The map is shifted from the original position.
   \\
7 &
  Goals &
  Purpose of the agent changes &
  * &
  * &
  * \\
8 &
  Processes &
  New type of state evolution not as a direct result of agent or entity action &
  * &
  A storm that impacts bird flight &
  Wind causes displacement of Steve
   \\ \bottomrule
\end{tabular}
\caption{Types of plausible novelties and their instantiations in our research domains. * denote that the environment doesn't support any instantiation.}
\label{tab:novelties}
\end{table}
\section{HYDRA}
\label{sec:hydra}

\begin{figure}
    \centering
    \includegraphics[width=1\textwidth]{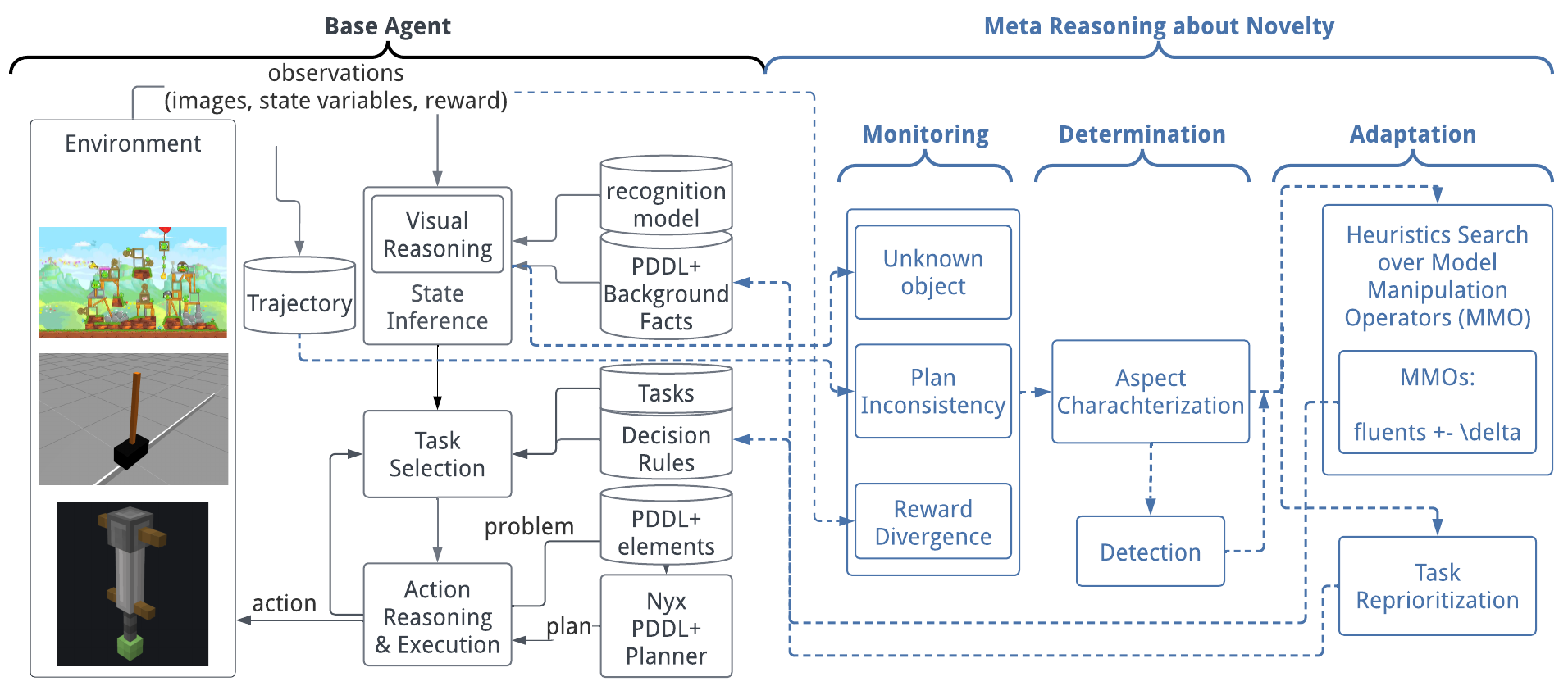}
    \caption{HYDRA - a domain-independent architecture for novelty reasoning}
    \label{fig:hydra_architecture}
\end{figure}

The main contribution of this work is the design of HYDRA, a domain-independent architecture for implementing a novelty-aware agent in complex, mixed discrete-continuous domains.  
The HYDRA architecture includes a \emph{base agent} and \emph{novelty meta-reasoning} components designed to detect novelties and adapt the base agent's behavior to them. A notional architecture is shown in Figure \ref{fig:hydra_architecture}. 
The base agent (Figure \ref{fig:hydra_architecture} - left, Section \ref{sec:base-agent}) implements a  \emph{perceive-decide-act} cycle in the environment. 
The novelty meta-reasoning components in HYDRA (Figure \ref{fig:hydra_architecture} - right, Section \ref{sec:novelty_reasoning}) detect the presence of novelty, characterize, and accommodate it by updating its knowledge bases as appropriate.

\subsection{Base Agent}
\label{sec:base-agent}
The base agent implements elements in the well-known Belief-Desire-Intention (BDI) theory~\citep{georgeff1999belief} that are similar to decision cycles implemented in prominent agent architectures: Soar \citep{laird2019soar} and ICARUS \citep{choi2018evolution}.
The main components of the HYDRA base agent are: 
(1) \emph{state inference}, which maintains and updates beliefs about the current state of the environment and augments them with background assumptions;
(2) \emph{task selection}, which selects the intermediate tasks the agent intends to perform;  
and (3) \emph{action reasoning and execution}, which uses a planner to determine the sequence of actions to execute in order to perform the selected task. \revision{Each component reasons about the current beliefs using a variety of long-term knowledge encoded in the agent (shown in cylinders in Figure \ref{fig:hydra_architecture}-left). }

HYDRA leverages PDDL+ planning to determine the sequence of actions to execute in order to perform the active task. A PDDL+ planning model defines the composition of the target system and its evolution via various \emph{happenings} that include actions the agent may perform, exogenous events that may be triggered, and durative processes that may be active. 
To support rich environments with complex dynamics and discrete and numeric state variables, we assume a planner that reasons over domains specified in PDDL+~\citep{fox2006modelling}. 

Recall that we consider an agent acting in a complex discrete-continuous environment $E$ that is modeled as a transition system $E{=}\tuple{S^*,A^*,H^*,F^*,G^*,O}$ where 
$S^*$ is a set of states of the environment;
$A^*$ is a set of the agent's actions; 
$H^*$ is a set of happenings that denote the environment's dynamics beyond the agent's control; 
$F^*$ is a transition function; $G^*$ is a set of goal states; and $O$ is a function that generates an observation given a state. 
A \emph{state} $s^* \in S^*$ of an environment is a complete assignment of values over all its propositional and real-valued variables $V^* = P^* \cup X^*$ at a given time $t(s^*) \in \mathbb{R}$.

\revision{To reason and plan in the environment, the autonomous agent (outlined in Algorithm \ref{alg:base_agent}) uses an internal model $D = (S, A, H, F, G)$ which is an approximation of environment $E$ and its components\footnote{\revision{Note that all components of model $D$ follow the definitions of the corresponding components of environment $E$.}}, i.e., $D \approx E$.}

\revision{The agent generates the initial state $s_0$ of its model given an observation $O_{init} = O(s_{obs} \subseteq s^*)$ from the environment and augmenting them with a set of background assumptions $B \approx s^* \setminus O_{init}$ about the obscured elements of the state $s^*$ of the environment $E$. Based on $s_0$, it selects a task $T$ that has a goal $G_T$. $G_T$ can either be a goal state in the environment ($G_T \in G*$) or an internal agent goal. HYDRA generated a planning problem $\mathcal{P}$ given the initial state $s_0$, and the goal $G_T$. HYDRA solves the resulting planning task, yielding a valid temporal plan $\pi$. The plan is then sequentially executed in the environment, and execution trace $\tau$ is collected for future analysis.}

\begin{algorithm}
\SetKwInOut{Input}{Input}\SetKwInOut{Output}{Output}
\Input{$O_{init} \subseteq s^*$, initial perception of the environment's state}
\Input{$R$, recognition model (domain ontology)}
\Input{$B_{PDDL+} \approx \{s^*\setminus O_{init}\}$, a set of background assumptions \\ about the state of the environment (in PDDL+)}
\Input{$\{T\}$, a set of tasks}
\Input{$D$, a PDDL+ domain of the environment}
\Output{$\tau$, plan execution trace (observations from environment)}
    $O_{PDDL+} \gets \emptyset$ \\ \label{alg1:top_line}
    \For{$o \in O_{init}$}{
        $O_{PDDL+} \gets O_{PDDL+} \cup \Call{TranslateToPredicateForm}{o, R}$ \\
        \Comment{map observations to domain ontology, translate to PDDL+}
    }
    $s_0 \gets \{B_{PDDL+} \cup O_{PDDL+}\}$ \Comment{Initial planning state $s_0 \approx s^*$} \\ 
    $G_T  \gets \Call{SelectTask}{\{T\}}$ \Comment{goal conditions $G_T$ for task $T$} \\ \label{alg1:line_task_selection}
    $\mathcal{P} \gets \Call{GenerateProblem}{s_0, G_T}$ \\ 
    $\pi \gets \Call{Plan}{\mathcal{P}, D}$ \Comment{solve planning task $\mathcal{P}$, generate plan $\pi$} \\ 
    \If{$\pi = \emptyset$}{
        \textbf{go to} \ref{alg1:top_line} \\ \Comment{Replan from new percepts $O_{init}'$, re-select task (\ref{alg1:line_task_selection}) if needed}
    }
    $\tau \gets \emptyset$ \\
    $O \gets s_0$ \\
    \For{$a \in \pi$}{
        $O' \gets \Call{ExecuteAction}{a}$ \Comment{collect observation after executing $a$} \\
        $\tau \gets \tau \cup (O, a, O')$ \\
        $O \gets O'$
    }
    \Return{$\tau$}
\caption{Base Agent Decision Cycle.}
\label{alg:base_agent}
\end{algorithm}

\subsubsection{State Inference}
The role of this HYDRA component (Algorithm \ref{alg:base_agent}, lines 1-4) is to describe the current state of the environment in sufficient detail to enable task selection and planning. It accepts observations $O_{init}$ about the current state from the environment and integrates it with a-priori knowledge or background assumptions about the domain (i.e., $B$ or $B_{PDDL+}$, the latter of which is expressed in PDDL+ but equivalent in content). \revision{Observations $O_{init}$ are a subset of the environment's state $s^*$, while the set of background assumptions $B \approx \{s^* \setminus O_{init}\}$ is the relative complement of $O_{init}$ w.r.t. the state of the environment $s^*$.} The resulting combination constitutes the initial planning state $s_0$ which is an approximation of the current true state of the environment $s^*$. Different environments provide a variety of input signals including visual information (e.g., images, object detection and colormap in ScienceBirds), continuous-valued sensor information (e.g., position and velocity of the cart and the pole in CartPole++), as well as discrete state information (e.g., locations and positions in PogoStick). State inference employs \emph{visual reasoning} to categorize known objects if visual information is provided as input (e.g., bounding polygon, a color map, etc.). As each domain may structure its inputs and outputs differently, the visual reasoning and state inference components may require domain-specific approaches, e.g., a dedicated recognition model $R$ to map detected objects to the defined domain ontology. 

\revision{Simply stated, the state inference component processes perception data (symbolic and/or visual) from the environment and automatically generates a set of PDDL+ facts that encode the agent's beliefs about the current state of the system/environment, its background knowledge and assumptions, as well as auxiliary state elaborations that are necessary to track task progress. The collated knowledge is then combined with the agent's goal specification and used to auto-generate a PDDL+ planning problem for the agent to solve.}

\begin{example}
In the ScienceBirds domain, the HYDRA accepts a set of visible objects on the scene including their locations and a vector representing their visual structures. The vector consists of the number of vertices in the bounding polygon, the area of the bounding polygon, as well as the a list of compressed 8-bit (RRRGGGBB) color and their percentage in the object. The HYDRA visual reasoning component for Science Birds accepts such vectors as input and categorizes the object as one of the known object classes using a recognition model. The recognition model is built with a standard implementation of multinomial logistic regression for multi-class classification and recognizes all known object types. 

The state inference component elaborates upon this symbolic information and adds current assumptions about the number of health points each pig and block have, gravity, starting velocity of the birds, how much damage a bird can incur about various objects and entities etc. A subset of PDDL+ facts generated for a specific initial state in ScienceBirds are below. 
\end{example}
\footnotesize
\begin{verbatim}
1. (:init 	(= (x_pig pig_-5932)  321) // input + visual reasoning
2. (= (y_pig pig_-5932)  20.0) // input + visual reasoning
3. (= (pig_radius pig_-5932)  4) // input + visual reasoning
4. (not (pig_dead pig_-5932) ) // elaboration
5. (= (pig_life pig_-5932)  10) // background assumption
6. (=	(bird_block_damage wood_-5912 wood_-5912) 0.5) // background assumption
7. (= (gravity)  9.8) // background assumption
...
\end{verbatim}
\normalsize

\subsubsection{Task Selection}
The main task of the HYDRA agent is based on the domain, such as, to craft a pogo-stick in PogoStick or kill all the pigs in the ScienceBirds domain. 
Planning directly toward performing this task is possible in some domains. 
In other domains, it is necessary or more efficient to progress toward this main task by setting intermediate tasks (also known as subtasks or subgoals)
for the agent to perform, and planning to achieve them. The HYDRA task selection component (Algorithm \ref{alg:base_agent}, line 5) implements this subtask selection mechanism, as follows.

Implementing a HYDRA agent for a given domain requires defining one or more HYDRA tasks. A \emph{HYDRA task} is a tuple $T=\tuple{\pre_T, D_T, G_T}$ 
representing the preconditions, domain, and goal of the task. 
The preconditions of a task ($\pre_T$) define when performing this task may be useful, and the goal of a task ($G_T$) defines what this task aims to achieve.  
The domain of a task ($D_T$) is a subset of the HYDRA agent domain $D$ specifies the parts of the domain relevant for performing the task. This is useful for efficiency reasons: performing some tasks do not require reasoning about all aspects of the domain. 

A HYDRA task is \emph{relevant} if its preconditions ($\pre_T$) are met and its goal ($G_T$) has not been achieved yet. Given, the current state and the set of tasks defined for the domain, the task selection components identify all relevant tasks (i.e., whose pre-conditions match the current state). From this set, it selects one based on a set of domain-specific prioritization rules and sets it the \emph{active task}. This is the task the HYDRA agent aims to plan for and act upon. There are certain conditions under which the active task is changed. If the base agent fails to find an executable plan or the found plan execution fails, HYDRA replans using newly obtained observations. If plan execution fails consistently and repeatedly, the HYDRA's task selection component may opt to choose a new active task, depending on domain-specific task prioritization rules.



\begin{example}
Task selection in non-novel environments for CartPole++ is trivial because there is only one task - maintaining the pole's upright angle. Task selection in Science Birds is straightforward; a task identifies which pig to shoot at and the domain-specific prioritization rules select the furthest pig (that can be shot with the lowest angle) first. A richer task selection component was implemented for PogoStick since the environment has additional agents that may interfere with the base agent's plans, 
and includes objects such as a safe or a chest, whose content is unknown and may have useful resources. 
Thus, the HYDRA agent for PogoStick includes exploratory tasks designed to gather information that might help craft the pogo-stick more efficiently.
Specifically, the tasks in our implementation of HYDRA for this domain include:
\begin{itemize}
    \item [$(T_1)$] Craft a pogo-stick
    \item [$(T_2)$] Interact with other agents.
    \item [$(T_3)$] Explore other rooms.
    \item [$(T_4)$] Open a safe or a chest.
    \item [$(T_5)$] Attempt to mine novel objects.
\end{itemize}
$T_1$ is the main task of the game. 
$T_2$ includes interacting with trader agents to learn the trade recipes they support. A trade recipe can be, for example, trading $9$ diamonds for $1$ titanium. This task is necessary because some items required for crafting the pogo-stick can only be obtained by trading with these trader agents. The domain of task $T_2$ is very limited, including only the location of the other agents and obstacles that may block getting to them, while the domain of task $T_1$ additional information such as crafting recipes. Tasks $T_1$ and $T_2$ are usually sufficient to solve the problem in non-novelty environments. Consequently, The task prioritization rules for this domain select $T_1$ first, followed by $T_2$. If after these tasks, the agent fails to craft a pogo-stick, tasks $T_3$, $T_4$, and $T_5$ are selected in a sequential order beginning at task $T_3$
\end{example}

\subsubsection{Planning and Execution}
\label{sec:planning}


 
In HYDRA, the agent's internal model $D$ is expressed in PDDL+ and it defines how the state changes under various actions, events that were triggered, and durative processes that are may be active\footnote{Functionally, the effects of applicable processes and events (i.e., encoded as $H \in D$) are implicitly modeled via a special \textit{time-passing} action $a_{tp}$ that the agent can apply alongside actions $A$ defined in the PDDL+ domain. In other words, the agent can wait for the environment to evolve before acting, as described in the theory of waiting \citep{mcdermott2003reasoning}. Any action in a plan $\pi$ can be a regular agent's action, or a time-passing action that advances the time and applies the effects of the environment's active processes and triggered events.}. Each happening is represented as a pair of \emph{preconditions} and \emph{effects} expressed in terms of assignments and mathematical expressions. An example of a process from the CartPole++ domain is shown in Figure \ref{fig:process-cartpole}.
A planning \emph{problem} $\mathcal{P}$ in the domain $D$ is a pair $\mathcal{P}=\tuple{s_0,G_T}$ where $s_0\in S$ is the initial state., and $G_T \subseteq S$ is a set of possible goal states in the current task $T$. A plan is a sequence of actions $\pi = (a_0, a_1, ..., a_n)$. A solution to a problem is a plan after execution reaches a goal state, i.e. $s_n \in G$, where $s_n$ is the state after executing action $a_n$ of plan $\pi$.
Executing a plan $\pi$ in domain $D$ yields a trajectory $\tau = (\tuple{s_0, a_0, s_1}, \tuple{s_1, a_1, s_2}, ... , \tuple{s_n, a_n, s_G})$, which comprises a sequence of tuples of the form $\tuple{s_i, a_i, s_{i+1}}$, representing that the agent performed action $a_i$ in state $s_i$, and reached state $s_{i+1}$.

PDDL+ planners accept a domain $D$ and a planning problem $\mathcal{P}$.
Thus, the first step in the HYDRA planning component is to encode the inferred current state and the goal of the active task as a PDDL+ problem 
and encode the domain of the active task as a PDDL+ domain. 
Due to the richness of the PDDL+ language, this encoding is relatively straightforward. 
The current state and task goal are represented as sets of grounded predicates and functions (discrete and numeric state variables). 
The domain encodes the definition of actions, events, and durative processes, each represented as a pair of \emph{preconditions} and \emph{effects} expressed in terms of logical and numeric conditions and assignments. 

\revision{HYDRA uses Nyx~\citep{piotrowski2024realworld}, a domain-independent PDDL+ planner to solve the generated planning tasks via heuristic search. 
} The output of a PDDL+ planner is either a solution to the given planning problem or a declared failure. If a plan is found, HYDRA executes it step by step in the environment. In case of a planning failure or repeated plan execution failures, the task selection component is called to reset the active task.

\begin{example}
In CartPole++, the output of the state inference component returns the velocity of the cart, the angle of the pole, etc. 
Each of these state variables is directly encoded in PDDL+ as numeric state variables (referred to as functions in PDDL+). 
The expected CartPole dynamics are encoded in the domain, which requires defining a process to specify the movement of the pole over time. 
An example of such a process is shown in Figure \ref{fig:process-cartpole}. 
\end{example}

PDDL+ defines deterministic models and dynamics, even though uncertainty is near-ubiquitous in real-world systems. Indeed, all application domains presented in this work are non-deterministic in some aspect, e.g., noisy object location data in Science Birds, randomized quantities of dropped resources in Polycraft, and drift in angular velocities in CartPole++. To account for environmental uncertainty, randomness, and non-determinism, HYDRA relies on replanning when necessary. As an example, when breaking down a tree in Polycraft, the number of spawned logs of wood is bounded random. The deterministic PDDL+ planning model of Polycraft always assumes the best-case scenario, i.e., that the maximum number of blocks is spawned. HYDRA replans if the agent collected an insufficient amount of wood logs. 


\begin{figure}
\centering	
\fontsize{8pt}{10pt}\selectfont
\begin{verbatim}
    (:process movement
    :parameters ()
    :precondition (and (ready)(not (total_failure)))
    :effect (and (increase (x) (* #t (x_dot)) )
    (increase (theta) (* #t (theta_dot)))
    (increase (x_dot) (* #t (x_ddot)) )
    (increase (theta_dot) (* #t (theta_ddot)) )
    (increase (elapsed_time) (* #t 1) ) ))
\end{verbatim}
\caption{Continuous PDDL+ process updating over time the positions and velocities of the cart and pole in CartPole++.}
\label{fig:process-cartpole}

\end{figure}

\subsection{Novelty Meta Reasoner}
\label{sec:novelty_reasoning}
HYDRA introduces a novel meta-reasoning process that implements the three steps of novelty reasoning in an open-world learning agent: detection, characterization, and accommodation. The process maintains explicit expectations about the agent's observations, transitions in the state space due to agent actions and extraneous dynamics, as well as its performance. A violation of these expectations indicates that a novelty has been introduced in the environment which must be further inspected, characterized, and accommodated for. Introducing such a process frames learning as a volitional activity undertaken by the agent to which resources are devoted only when an opportunity presents itself (as indicated by violation of expectations). This is in stark contrast with the classical machine learning setup in which agent learning is controlled externally. 

Specifically, HYDRA (Figure \ref{fig:hydra_architecture}) implements (1) a set of \emph{novelty monitors} that maintain a variety of explicit expectations about the environment evolution and monitor divergence; 
(2) a \emph{novelty determination} component that aggregates the information from the monitors and determines if a novelty has been introduced and requires adaptation; and (3) adaptation strategies: a \emph{heuristic search-based model repair} component that manipulates the base agent's PDDL+ model to be consistent with the detected novelty and a \emph{task re-prioritization component}. 

\subsubsection{Novelty Monitors}
Novelty monitors are a set of components that maintain explicit expectations about various aspects of agent behavior in the environment. They capture violations of those expectations which are processed by further downstream reasoning. Monitors are dedicated to different parts in the base agent reasoning cycle and are implemented using diverse computational techniques.  
\paragraph{Unknown Objects and Entities}
This monitor encodes an expectation that all observed entities processed by the state inference module are \emph{known} i.e., they are of a type that is encoded in the agent's domain ontology and its planning model. A recognition model $R$ processes input observations and maps them to known entities. If an object appears in the environment whose type cannot be recognized using the recognition model or is not in the agent's planning model, the monitor flags the existence of novelty. In some domains the recognition model is trivial and this monitor is a binary operation: either a new entity type has been observed or not. This is the case in the PogoStick domain, where the Polycraft environment provides the type of each object which can be matched against a type inventory. In domains with visual input, such as ScienceBirds, a more nuanced recognition model is leveraged. The recognition model (Section \ref{sec:base-agent}) has been trained for near-perfect performance in canonical cases - i.e., it can categorize objects with high confidence (prediction probability $\geq threshold_c$). If the recognition model produces detections with low confidence (prediction probability $< threshold_c$), the monitor flags the likelihood of a novelty.


\paragraph{Plan Inconsistency}
Let $D$ be the planning domain used by the HYDRA planning component. 
The second novelty monitor, referred to as the \emph{planning domain inconsistency} monitor, measures how accurately $D$ describes the environment dynamics, i.e., the behavior of the different happenings (actions, effects, and processes). 
This novelty monitor is thus geared towards detecting novelties that change these environment transitions. 
Specifically, this monitor relies on computing an \emph{inconsistency score}, denoted $C(\pi, D, \tau)$ where $\pi$ is a plan, $D$ is a planning domain, and $\tau$ is the trajectory observed when performing $\pi$ starting from state $s$. The inconsistency score $C(\pi, D, \tau)$ quantifies the difference between the observed trajectory $\tau$ and the trajectory we expected to observe when performing $\pi$ in $s$ according to $D$. This expected trajectory can be obtained by simulating the plan $\pi$ in $s$ according to $D$. This is possible since $D$ specifies the expected environment dynamics, including actions' effects. 
Existing tools such as VAL~\citep{howey2004val} support such simulations although we wrote our own domain-independent PDDL+ simulator. 
We computed the inconsistency score $C$ as the ``distance'' between pairs of corresponding states in observed and simulated trajectories, discounted proportionally by time to give more weight to changes that are observed earlier in the trajectories.

There are many ways to measure this ``distance'', and the exact function can be domain-dependent. 
An example of this distance measure is the Euclidean distance between the pairs of corresponding states in the observed and expected trajectories. 
In the most general case, the distance would be calculated over all state variables $V$ (where propositional values are cast as 1 or 0). 
However, for increased accuracy, this can be restricted to a subset of relevant state variables.
Formally, let $S(\tau)$ be the sequence of states in the observed trajectory and $S(\pi,D)$ be the expected sequence of states obtained by simulating the generated plan $\pi$ with respect to the domain $D$. Let $S(x)[i]$ denote the $i^{th}$ state in the state sequence $S(x)$ \revision{where $i \leq |\tau|$ and $i \in \mathbb{Z}_{\geq 0}$}. Inconsistency score of domain $D$ is computed as:

\begin{equation}
    C(\pi, D, \tau) = \frac{1}{|\tau|} \sum_{i} \gamma^i\cdot ||S(\tau)[i] - S(\pi,D)[i]||
\end{equation}
where $0<\gamma<1$ is the discount factor. Errors in $C$ due to sensing noise, rounding errors, and other issues can accumulate over time. 
Consequently, the Euclidean distance between corresponding states is likely to be higher later in the trajectories. 
The discount factor $\gamma$ prevents such errors from dominating the inconsistency score.

In a non-novel environment $E$ with an ideal domain $D$, the expected evolution of the system predicted by the planner should perfectly match the observed behavior in simulation, i.e., the two resulting trajectories align by default with an inconsistency score $C=0.0$. 
In complex environments, however, this is usually impossible to achieve due to rounding errors, perception inaccuracies, and similar common issues. 
To account for such noise, the planning domain inconsistency monitor relies on a domain-specific inconsistency threshold $C_{th}$, where inconsistency scores below this threshold are ignored. 
Setting $C_{th}$ requires striking a fine balance between accurately estimating the inconsistency score and suppressing noise stemming from the execution environment. \revision{Currently, the inconsistency threshold $C_{th}$ is set manually for each environment by the planning model designer, based on their knowledge, to reflect the accuracy of the PDDL+ domain.}

\begin{figure}[t]
	\centering
	\includegraphics[width=0.8\columnwidth]{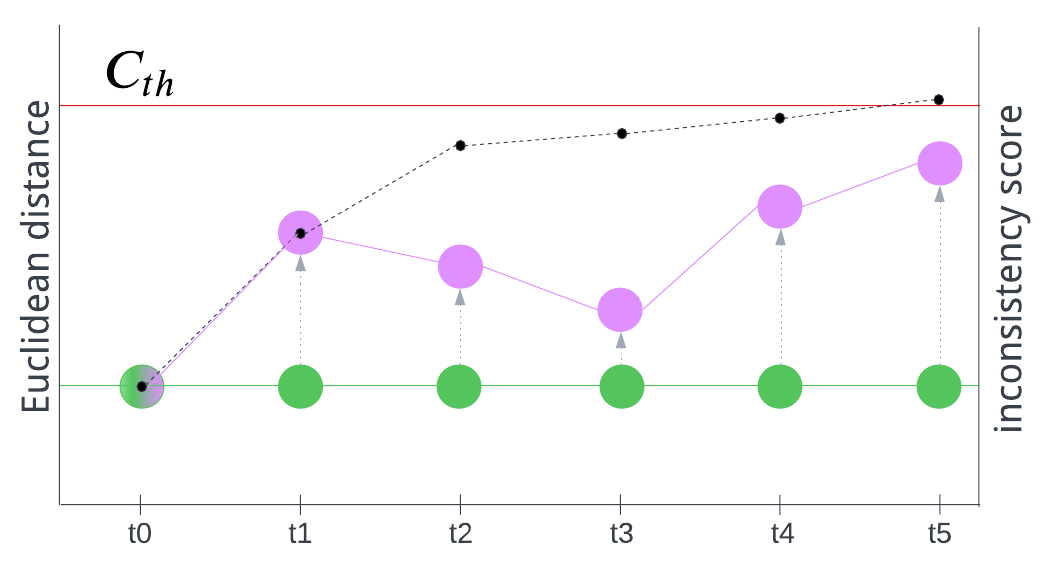}
	\caption{Visualization of the inconsistency score (black, dashed line) of the expected state trajectory obtained from the plan generated with agent's internal model $D$ (pink nodes) with respect to the observations trajectory $\tau$ from the environment $E$ (green nodes).}
	\label{fig:inconsistency_trajectory}
\end{figure}

A graphical representation of state trajectory-based inconsistency score computation is shown in Figure~\ref{fig:inconsistency_trajectory} in which the expected state trajectory under the agent's internal model $D$ (pink nodes) is compared against the observed trajectory (green nodes) in the environment $E$. The left y-axis denotes the Euclidean distance between states and the right y-axis is the inconsistency score. Due to the discount factor $\gamma$, the distance between states later in the trajectory contributes less to the inconsistency score than the distance between states earlier in the trajectory. The planning domain inconsistency monitor returns a non-zero inconsistency score if the inconsistency threshold is exceeded.

\begin{example}
\label{sec:ex:inconsistency}
CartPole++ domain is simple with a few relevant elements (pole, cart, etc.) and our PDDL+ model is fairly accurate. Consequently, we use the Euclidean distance between states as our inconsistency score and set the threshold to a very low $0.009$. In contrast, the ScienceBirds domain is extremely complex with several objects and entities that are modeled in PDDL+ with varying levels of fidelity. The inconsistency score for ScienceBirds is engineered to focus on information that is relevant for good gameplay. Specifically, we record if there is a mismatch ($m_i$) in pigs between plan simulations and observations i.e., they exist in the plan but are dead in observations and vice-versa. Next, we measure the difference ($\Delta h$) between the maximum height achieved by a bird in plan simulation and observations. The inconsistency score is computed as $\sum_i 50 \times m_i + \Delta h_i$ where $i$ is a shot. $C_{th}$ is set high at $10$ to alleviate inaccuracies in PDDL+ modeling.
In PogoStick, inconsistency is measured as the difference in the number of objects in the simulated and observed traces. Difference of other properties (such as location etc.) is weighted and added to the sum to give the final inconsistency score. The threshold $C_{th}$ is set to $2$, i.e. novelty is detected and repair is called if the count of object difference is greater than $2$.
\end{example}

\paragraph{Reward Divergence} \label{sec:reward_divergence}
\begin{figure}
\begin{center}
\includegraphics[width=13.5cm]{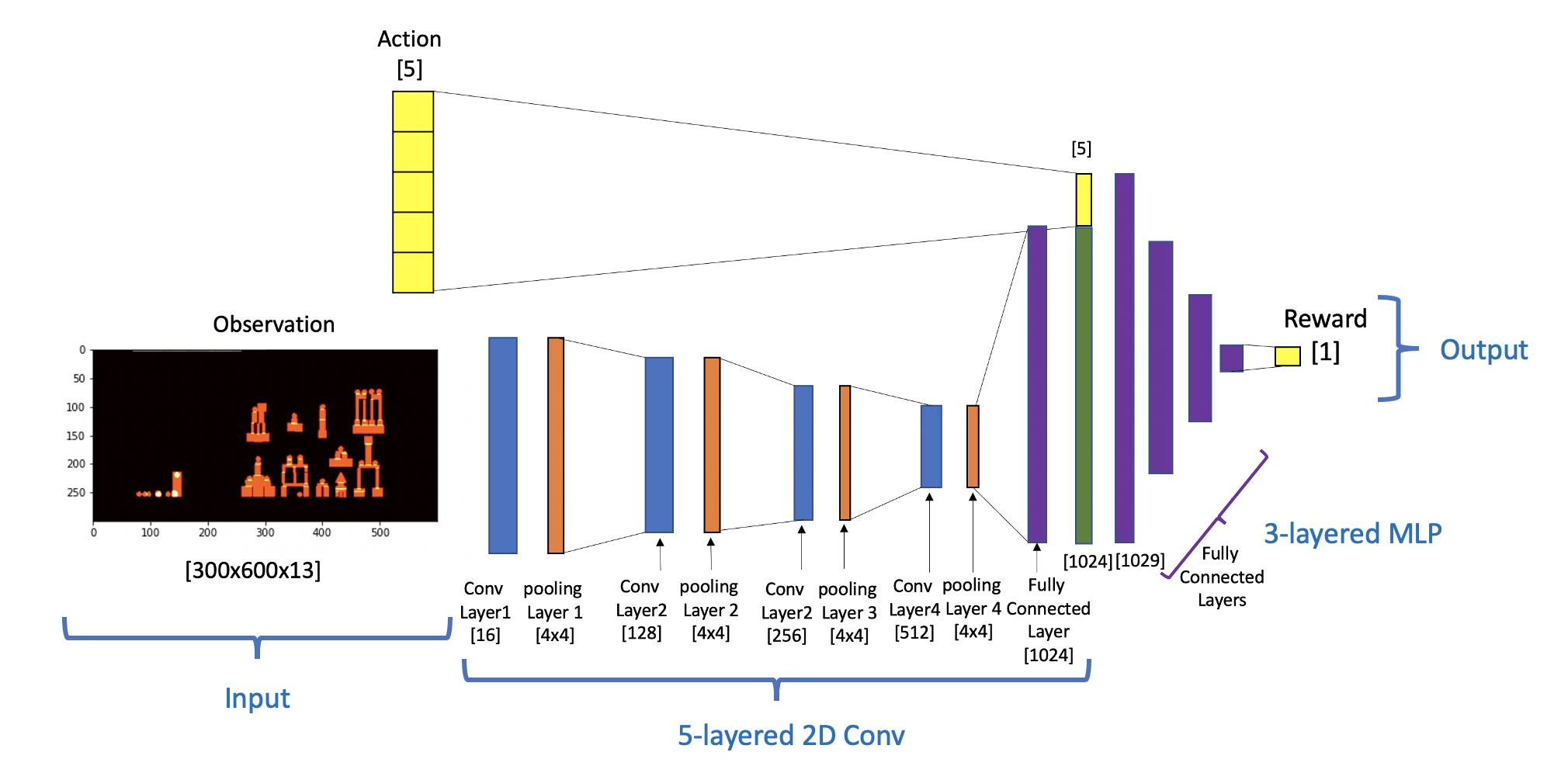}
\end{center}
\caption{Neural network architecture of the reward estimator for ScienceBirds}
\label{fig:reward_estimator}
\end{figure}
Our third novelty monitor maintains expectations about the quality of its own performance. Any change in performance quality can indicate that a novelty has been introduced in the environment. We leverage the reward signal generated by environments such as ScienceBirds to gauge changes in performance quality. 

Intuitively, assume $r(s,a)$ to be the reward collected by the agent when performing action $a$ in state $s$ in the non-novel environment. After a novelty $\nu$ is introduced in the environment, a different reward $r_\nu(s,a)$ is accumulated. Reward divergence is the absolute difference between the expected reward (given prior experience in non-novel environments) and the reward received upon performing an action. The difference is higher if $\nu$ impacts the agent's performance. 

We developed a neural-network-based reward estimator that serves as a surrogate, denoted as $g(s,a)$, that returns an estimate of $r(s,a)$ in canonical settings. 
We train this estimator by performing actions in the non-novel environment and training the neural network to minimize the root mean squared loss ($\mathcal{L}$) between $g(s,a)$ and $r(s,a)$ through the following optimization problem:
\begin{equation}
 \mathcal{L}= \min_{\Theta} \sqrt{\frac{1}{N} \sum_{i=1}^{N}  \|g(s_i,a_i)-r(s_i,a_i)\|_2^2}
\end{equation}
Where $N$ is the number of training data points and $\Theta$ is the set of parameters in our estimator $g$. 
Given the estimator $g$ pre-trained with non-novel data, we implemented the reward divergence novelty monitor by computing the absolute error between the predicted reward from $g$ and the ground truth reward $r$, computed as follows:
\begin{equation}
 R_{div}(s, a, r) =  \|g(s, a) - r(s, a)\|,
\end{equation}
where $s$, $a$, and $r(s,a)$ are the observed state, executed action, and reward collected in a possibly novel environment. The magnitude of this estimated reward divergence score, $R_{div}$, serves as an indicator of how much the reward deviated in an environment with a novel feature as compared to a non-novel environment, given identical actions taken in the same state. Thus, a larger value of $R_{div}$ implies a more pronounced deviation in reward, suggesting that novelty has been introduced.

\begin{example}
We implemented the reward divergence novelty monitor for the ScienceBirds domain, as follows. Since a state in ScienceBirds can be represented as a visual scene composed of multiple channels of different objects, we utilize Convolutional Neural Networks (CNNs) for our reward estimator ($g$), which have been proven to be successful in learning visual representations. The architecture of our reward estimator for ScienceBirds is illustrated in Figure~\ref{fig:reward_estimator}. 
Our reward estimator is designed to receive a pairing of observational state and action as input to predict reward as output. The architecture of the estimator is comprised of four convolutional layers, followed by four fully connected layers, incorporating a Rectified Linear Unit (RELU) activation function. The convolutional layers extract features from the observation, which are then transformed into a flattened representation and concatenated with the action. Then, fully connected layers subsequently predict a scalar reward given this concatenated representation of action and observation features. 
\end{example}

\subsubsection{Novelty Determination}
Each novelty monitor generates information about the existence of novelty based on various aspects of agent behavior (unknown objects in the observation space, environment transitions, and performance quality). The novelty determination component collects information from all novelty monitors as well as some other observations about the agent (e.g., success or failure at the overall task) and determines if novelty has been introduced and the agent's domain needs to be adapted.  We have explored several approaches for this component, but eventually implemented domain-specific decision rules. Table \ref{tab:monitors} summarizes novelty detectors built for different domains.

\begin{example}

For the CartPole++ agent, only the plan inconsistency novelty detector was used as the PDDL+ domain accurately reflected all of the relevant characteristics of the environment. Upon detecting novelty, the planning model is adapted (described in Section \ref{sec:accomodation}).

For the ScienceBirds environment, we declare that novelty if (a) either there is an unknown object or (b) if at least one of the other two novelty monitors - plan inconsistency or reward divergence - exceeds their threshold for more than $3$ consecutive episodes. If there is an unknown object, a bird is shot at it. If either of the two monitors has detected novelty, decision-making in HYDRA is adapted (described in Section \ref{sec:accomodation} and Section \ref{sec:reprioritization}).

For PogoStick, we infer the existence of novelty based on whether (a) the scene contains unknown objects or (b) the plan inconsistency exceeds the set threshold. In the first case, the object is picked up by the agent character Steve, and in the latter case, the plan model is adapted (Section \ref{sec:reprioritization}).
\end{example}

\begin{table}[ht]
\centering
\begin{tabular}{@{}p{2.45cm}p{10.8cm}@{}}
\toprule
Domain & Novelty detectors \\
\midrule
\multirow{3}{*}{PogoStick} & 
\tabitem Unknown objects or entities in the environment. \\
& \tabitem PDDL+-based tracking of agent's inventory contents.\\\midrule
\multirow{2}{*}{Cartpole++} & \tabitem PDDL+-based tracking of the positions and velocities of the cart and pole. \\ \midrule
\multirow{4}{*}{Science Birds} & 
\tabitem Unknown objects or entities in the environment. \\
& \tabitem Reward prediction. \\
& \tabitem PDDL+-based tracking of target survivability and tracking of bird height \\
\bottomrule
\end{tabular}
\caption{Novelty monitors designed for each domain.}
\label{tab:monitors}
\end{table}

\subsubsection{Accommodation through Heuristic Search-Based Repair}
\label{sec:accomodation}

The proposed search-based model repair algorithm works by searching for a \emph{domain repair} $\varPhi$, which is a sequence of model modifications that, when applied to the agent's internal domain $D$, returns a domain $D'$ that is consistent with the observed trajectories. \revision{Formally, novelty accommodation is a model adaptation process $\textrm{Repair}(\varPhi, D) \rightarrow D'$, such that $D' \approx E'$. In other words, the novelty accommodation process updates model $D$ to be consistent with the novel environment $E'$. }
\smallskip

To find such a domain repair, our algorithm accepts as input a set of possible basic \emph{Model Manipulation Operators} (MMOs), denoted $\{\varphi\} = \{\varphi_0, \varphi_1, ... , \varphi_n\}$. Each MMO $\varphi_i \in \{\varphi\}$ represents a possible change to the domain. Thus, a domain repair $\varPhi$ is a sequence of one or more basic MMO $\varphi_i \in \{\varphi\}$. An example of an MMO is to add a fixed amount $\Delta\in\mathbb{R}$ to one of the numeric domain fluents.
In general, one can define such an MMO for every domain fluent. In practice, however, not all domain fluents are equal in their importance thus the repair can be focused on a subset of state variables that the domain designer deems relevant \revision{(designated ``repairable fluents")}.

\begin{algorithm}
\SetKwInOut{Input}{Input}\SetKwInOut{Output}{Output}
\Input{$\{\varphi\}$, a set of basic MMOs}
\Input{$D$, the original PDDL+ domain}
\Input{$\pi$, plan generated using $D$}
\Input{$\tau$, a trajectory}
\Input{$C_{th}$, inconsistency threshold}
\Output{$\varPhi_{best}$, a domain repair for $D$}
OPEN$\gets\{\emptyset\}$; 
$C_{best}\gets\infty$;
$\varphi_{best}\gets\emptyset$\\
\While{$C_{best}\geq C_{th}$}{
    $\varPhi\gets$ pop from OPEN\\
    \ForEach{$\varphi_i\in\{\varphi\}$}{
        $\varPhi'\gets \varPhi \cup \varphi_i$ \Comment{Compose a domain repair} \\
        $D' \gets$ Repair($\varPhi'$, $D$) \\
        $C_{\varPhi'} \gets$ InconsistencyEstimator($\pi$, $D'$, $\tau$)\\
        \If{$C_\varPhi \leq C_{best}$}{
            $C_{best} \gets C_{\varPhi'}$\\
            $\varPhi_{best} \gets \varPhi'$
        }
        Insert $\varPhi'$ to OPEN with key $h(\varPhi', C_{\varPhi'})$ \nllabel{alg:line:f} \\
    }
}
\Return $\varPhi_{best}$\\
\caption{PDDL+ general model repair algorithm.}
\label{alg:repair}
\end{algorithm}

Algorithm~\ref{alg:repair} lists the pseudo-code for our search-based model repair algorithm. 
Initially, the open list ($OPEN$) includes a single node representing the empty repair, 
and the best repair seen so far $\varPhi_{best}$ is initialized to an empty sequence. 
This corresponds to not repairing the agent's internal domain at all. 
Then, in every iteration, the best repair in OPEN is popped, and we compose new repairs by adding a single MMO to this repair, and add them to $OPEN$.  
For every such repair $\varPhi'$ we compute an inconsistency score $C_{\varPhi'}$. 
This is done by modifying the agent's internal domain $D$ with repair $\varPhi'$, simulating the actions in plan $\pi$, and measuring the difference between the simulated outcome of these actions and the observed trajectory $\tau$. 
The inconsistency score $C_{\varPhi'}$ serves two purposes. First, we keep track of the best repair generated so far, and return it when the search halts. Second, we consider a repair's inconsistency score when choosing which repair to pop from $OPEN$ in each iteration. This is embodied in the heuristic function $h(\varPhi', C_{\varPhi'})$ in line~\ref{alg:line:f} in Algorithm~\ref{alg:repair}. 
In our implementation, $h$ is a linear combination of the inconsistency score and the size of the repair, i.e., the number of MMOs it is composed of. The latter consideration biases the search towards simpler repairs. 

\begin{figure}
	\centering
	\includegraphics[width=0.9\columnwidth]{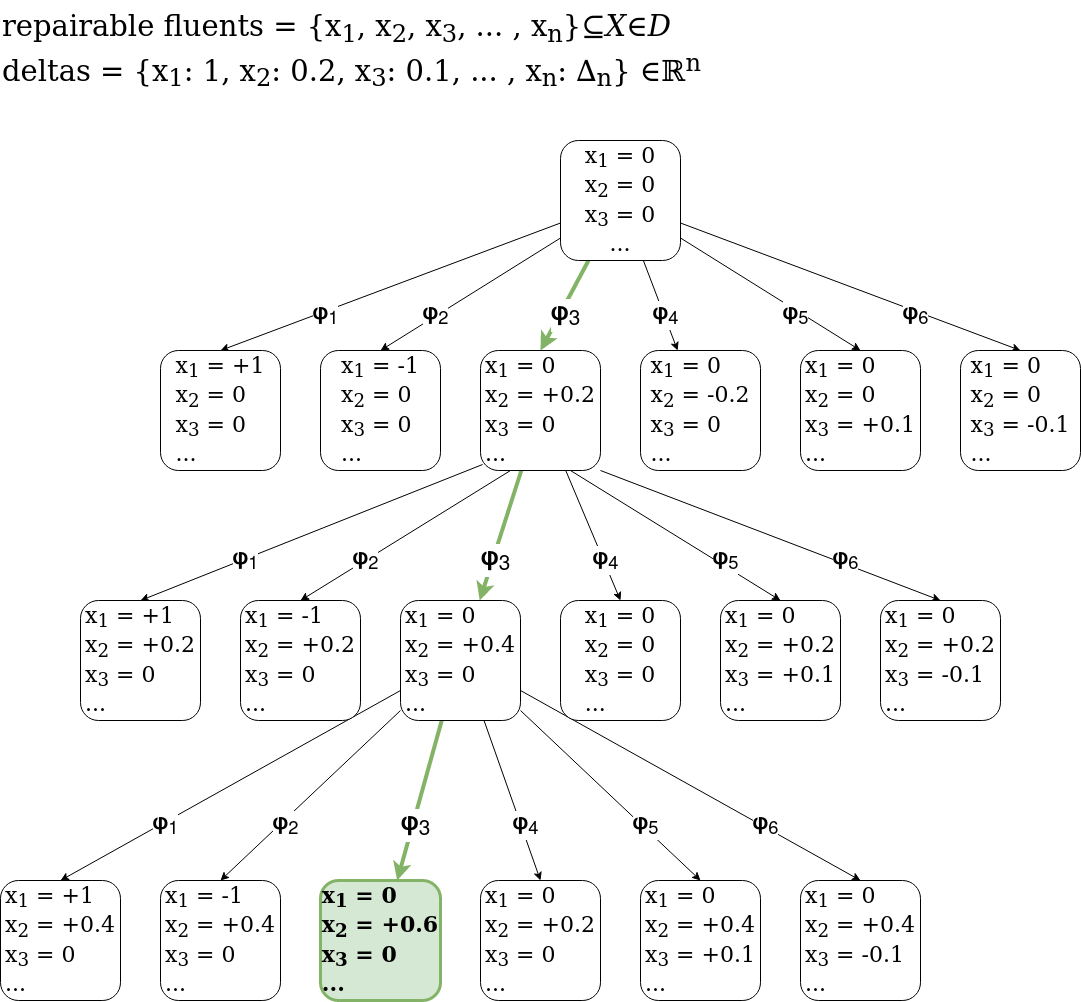}
	\caption{An example MMO search graph showing how a repair (a sequence of MMOs) is selected. The green node depicts a goal state in the MMO repair search, the minimal change to the model that causes the inconsistency score to fall below the set threshold.}
	\label{fig:mmo_search}
 \vspace{-3mm}
\end{figure}

Figure~\ref{fig:mmo_search} visualizes an example search tree of the MMO-based model repair algorithm where MMOs are treated as actions and the state is composed of changes to the default model (0 indicates no change to the given fluent). The inconsistency score is estimated for each generated repair, and the search terminates once a repair is found such that the updated domain $D'$ is consistent with the true transition function $F^*$. MMO repair is performed on a set of repairable fluents $\{x_1, x_2, ..., x_n\}$. Each MMO adjusts the value of a given fluent by a fixed amount $\Delta$ ($+\Delta$ or $-\Delta$), defined \emph{a priori} per fluent (denoted as `delta' in the top left of Figure.~\ref{fig:mmo_search}). In this example scenario, the best repair $\varPhi_{best} = \{\varphi_3, \varphi_3, \varphi_3\}$ is a sequence of three MMOs (each adjusting $x_2$ by $+0.2$) such that repair $\varphi_{best}$ changes a single state variable $x_2 \in X$ by adding $0.6$ to its value.

\paragraph{Focused Model Repair}

In many cases, allowing the repair to adjust multiple variables simultaneously can result in a state space explosion. The branching factor of the general model repair algorithm is $2^n$. An impactful novelty usually requires significant change to the domain, high branching factor, and vast search space might cause it to not be found within a reasonable time.

\begin{figure}
	\centering
	\includegraphics[width=0.9\columnwidth]{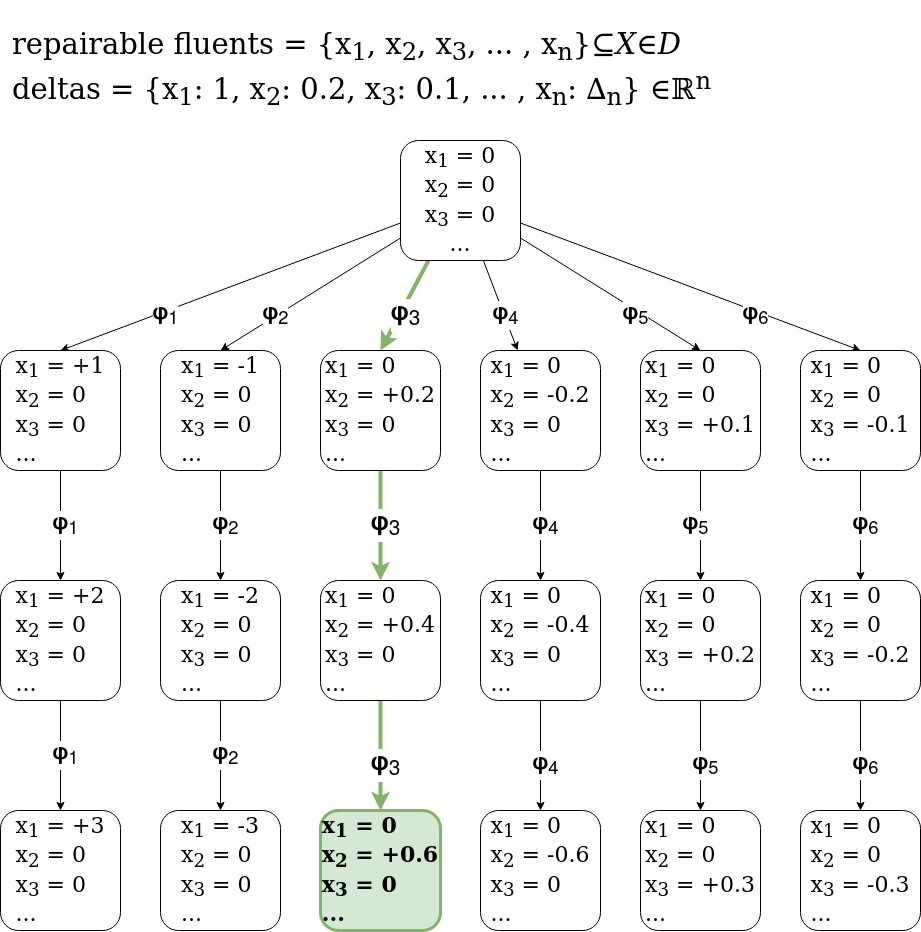}
	\caption{Example MMO search graph for a focused model repair.}
	\label{fig:mmo_focused_search}
 \vspace{-4mm}
\end{figure}

To make repair feasible in complex domains, we introduce focused model repair, a restricted variant of the general algorithm. The two mechanisms differ only in how they expand new repair candidates. In the focused case, an additional constraint is imposed such that any repair candidate $\varPhi$ can only contain MMOs of the same type $\varPhi = \{\varphi_i, \varphi_i, ... , \varphi_i\}$ where $\varphi_i \in \{\varphi\}$. 
To implement focused repair, the aforementioned constraint is added after line 4 in Algorithm \ref{alg:repair}. The search graph of the focused model repair is shown in Figure \ref{fig:mmo_focused_search}.

\revision{The focus of this work is reasoning with \emph{single persistent novelty} which assumes that any domain shift has a single root cause, even though its effects can permeate throughout the system and alter multiple sensor readings. For example, consider a mobile robot moving on a grid in cardinal directions. If the robot is rotated slightly, both its X and Y coordinates will be erroneous. However, the cause of the unexpected inaccuracy can be traced to a single change such as a novel surface with reduced/increased friction, worn-out robot wheels, or lower-than-expected torque. HYDRA repairs the model by targeting the root cause of the novelty. As highlighted in Section~\ref{sec:problem}, in real-world systems, single-fault scenarios are by far the most probable. We do not consider cascading faults where new separate faults develop as a consequence of previous faults.}


\subsubsection{Accommodation through Task Re-prioritization}
\label{sec:reprioritization}
The second adaptation strategy implemented in \textsc{HYDRA} is changing the order in which tasks are solved. The base agent (in Figure \ref{fig:hydra_architecture}) selects an active task from all relevant tasks based on domain-specific decision rules. The decision rules are designed based on how various tasks are related to each other in non-novel environments. Under certain novelties, these pre-determined task prioritization decision rules can become unproductive. These novelties are \emph{obstructive}; successfully achieving a task higher in the prioritization order obstructs achieving a task later in the order. HYDRA accommodates for such novelties by changing the order in which it attempts tasks.

Current accommodation implementation is a blind re-prioritization strategy. The task at the top of the priority list is removed and added at the end. A more sophisticated method will prioritize the tasks based on the expected value of attempting each task order, which is estimated based on its successes and failures in novel conditions.

\begin{figure}
    \centering
    \includegraphics[width=0.45\textwidth]{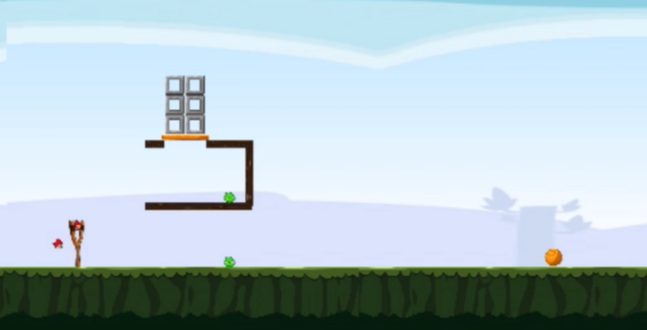}
    \includegraphics[width=0.45\textwidth]{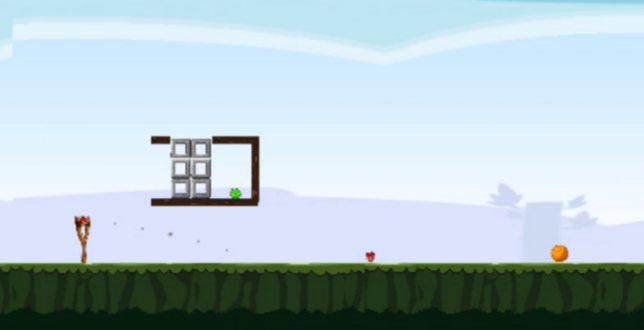}
    \caption{Example of an obstructive novelty. Left shows the beginning of the game in which the wooden platform is holding a tower of blocks. Right shows the consequence of hitting the pig at the bottom; the wooden platform collapses making the game unsolvable.}
    \label{fig:goal_relation}
\end{figure}

\begin{example}
Tasks in ScienceBirds select which pig to aim at and decision rules sequence the pigs based on the shot angle they can be shot at, with the pig that can be shot at the smaller angle first. With this task prioritization scheme, certain novel Science Bird games are unsolvable. An example is shown in Figure \ref{fig:goal_relation}. The novelty in this scenario is the introduction of a destructible wooden platform (left) that collapses as soon as the bottom pig is hit. The blocks held by the platform obstruct any shots at the top pig (right), making this game unsolvable. HYDRA accommodates by inverting the order in which pigs are shot; which is a successful strategy and is maintained.
\end{example}

\section{Evaluation}
\label{sec:evaluation}
Here we evaluate the efficacy of various novelty meta-reasoning methods implemented in supporting online detection, characterization, and accommodation of novelties. The results presented in this section have been generated using the novelties that have been bundled with our research domains - CartPole++, ScienceBirds, and PogoStick and are publicly available. \revision{A dataset containing PDDL+ planning domains and non-novel problems for each presented environment can be found here: \url{https://zenodo.org/record/8417802}}

\subsection{Novelty Detection}
First, we study the sensitivity of the three implemented monitors -- unknown objects and entities, plan inconsistency, and reward divergence -- to the space of novelties available with our research domains. The three domains provide different types of input to the agent and consequently, a subset of monitors was implemented for each domain. \revision{Inconsistency thresholds were selected per domain by analyzing inconsistency measurements in non-novel trials. Selecting the inconsistency threshold via experimentation requires a balance between minimizing the number of false positive reports (ensuring that noise is not flagged as novelty) and minimizing false negatives (ensuring that novelty is not misinterpreted as noisy measurements).}
The plan inconsistency monitor is tied closely with the heuristic search-based model repair method for accommodation. Consequently, it was evaluated with the efficacy of repair, and the results are summarized in Section \ref{sec:evaluation:accomodation}. 

\subsubsection{Unknown Objects and Entities}
This monitor was implemented only for PogoStick and ScienceBirds.
The monitor was trivial to develop for PogoStick because the input to the agent is symbolic and each object is accompanied by its type. The agent implements a typed inventory and can match the label against known types to detect novelties. 
ScienceBirds provides a colormap for each object on the scene which requires visual reasoning to infer object types.
\paragraph{Training} The recognition model for ScienceBirds was built using a standard implementation of multiclass (one-versus-rest) logistic regression\footnote{\href{https://scikit-learn.org/stable/modules/generated/sklearn.linear\_model.LogisticRegression.html\#sklearn.linear\_model.LogisticRegression}{sklearn.LinearModel.LogisticRegression}}.
It was trained on a dataset of $13,198$ datapoints obtained by sampling ScienceBirds non-novel levels using a standard $80\%/20\%$ train-test split with L2 regularization. The recognition model could recognize the $13$ known object types with $100\%$ accuracy with the confidence threshold of $0.65$. 
\paragraph{Analysis} The monitor was tested on $6$ types of ScienceBirds levels (covering novelty ID 2 in Table \ref{tab:novelties}) that introduced news objects or entities. Each level type was sampled $10$ times to create a novelty recognition test set and the ScienceBirds agent played these levels. The monitor could detect novelty in $6$ level types with $100\%$ accuracy. One novelty level introduced a new orange-colored bird entity. This bird was detected as a red bird - a known object type. While the novelty was not reported, the agent could still play the game. 

Arguably, the recognition module implemented here is simple. A more complex visual input (pixels) would motivate a complex recognition module built with convolutional neural networks. All types of classifiers estimate class probabilities during classification which can be used with an appropriate threshold to trigger novelty detection.

\subsubsection{Reward Divergence}
In order to demonstrate the utility of the reward divergence monitor introduced in Section~\ref{sec:reward_divergence}, we evaluate the sensitivity of this score to the existence of novelties in ScienceBirds. 

\begin{figure}
    \centering
    \includegraphics[width=0.7\textwidth]{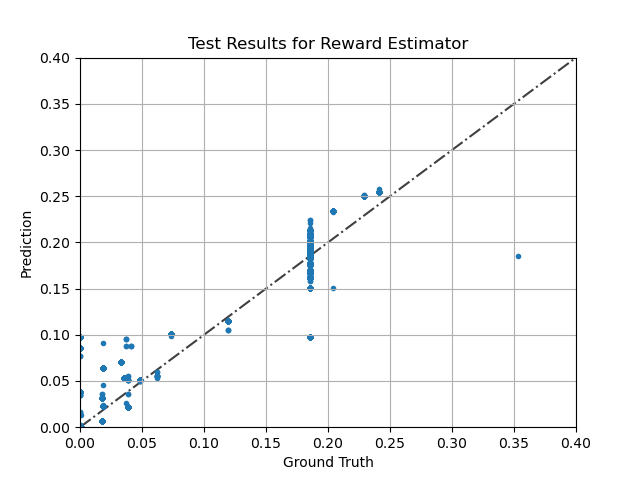}
    \caption{Scatter plot of reward predicted by the reward estimator against ground truth reward generated by ScienceBirds. The points are scattered around the $45$-degree line indicating highly accurate predictions.}
    \label{fig:reward_estimator_scatter_plot}
\end{figure}

\paragraph{Training} We first trained the reward estimation model ($g*$) on approximately $15,000$ data collected from non-novel ScienceBirds levels. We used 80$\%$ for training and 20$\%$ for testing. To train the reward estimator, we normalized the scale of the input (state and action) and output (reward) to fall within the range $(0,1)$. Scatter plot of estimated and actual rewards in the test set are shown in Figure \ref{fig:reward_estimator_scatter_plot}. The points aggregate close to the $45$-degree line. The root mean square error (RMSE) score of the fully-trained reward estimator ($g^*$) tested in the non-novel environment was $0.008$, indicating that the reward estimator can accurately estimate rewards for actions taken in non-novel levels. 

\begin{figure*}
    \includegraphics[width=\textwidth]{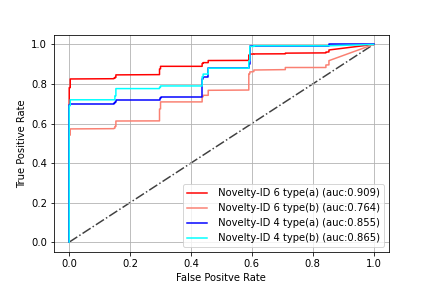}
    \caption{ROC curves for various Novelty ID instantiations in ScienceBirds. $0\%$ false positive rate is achieved at $2.96$, $2.97$, $1.86$, $2.15$ thresholds for Novelty ID 6 (a), (b), and Novelty ID 4 (a), (b) respectively.}
    \label{fig:sb_roc}
\end{figure*}

\paragraph{Analysis}
To evaluate the sensitivity of reward divergence, we created novelty test datasets for all published novelties in the ScienceBirds domain. Each dataset was sampled from levels instantiating a specific novelty ID and non-novel levels and consisted of datapoints with state, action, reward tuples. We used the reward estimator trained on non-novel levels ($g^*$) to estimate reward for each datapoint and computed absolute reward error score. We then analyzed if the absolute error in reward estimation is sensitive to the existence of novelties and can be used to classify a datapoint as novel. Table \ref{tab:sb_auc} shows the area under the receiver operating characteristic (ROC) curve (AUC) for various novelty IDs in the ScienceBird domain.  The results suggest that, indeed, absolute error in reward estimation is sensitive to existence of some types of novelties but not all. Specifically, it can detect novelties belonging to IDs 4 (interaction) and 6 (global constraints). While new types of relational interaction between objects and entities can impact the cumulative score the agent gets, changes in global constraints can lead to significant plan execution failures. ROC curves in Figure \ref{fig:sb_roc} suggest that for threshold value specific novelty IDs can be detected with high accuracy using the threshold value of $2.0$. 

\begin{table}[ht]
\begin{tabularx}{\columnwidth}{lll}
\toprule
Novelty ID & Type  & AUC  \\ \midrule
2           & (a) New styrofoam block     & 0.536  \\
2   & (b) New orange bird & 0.536 \\
2             & (c) New entity that teleports birds     &  0.530 \\
4 & (a) Magician turns objects into pigs & \textbf{0.855} \\
4 & (b) Butterfly adds health points to the pigs nearby & \textbf{0.865} \\
5            & Blue bird splits into 5    & 0.546  \\
6       & (a) Gravity stops impacting birds    & \textbf{0.909}  \\
6       & (b) A radioactive area that kills any bird     & \textbf{0.764} \\
8      & A time-triggered storm that applies force on birds     & 0.528  \\
                            \bottomrule
\end{tabularx}
\caption{Area Under the ROC Curve (AUC) of the reward divergence-based novelty monitor tested with ScienceBirds novelty levels instantiating various types of Novelty IDs}
\label{tab:sb_auc}
\vspace{-3mm}
\end{table}


\subsection{Novelty Accommodation}
\label{sec:evaluation:accomodation}
Novelty accommodation in HYDRA was evaluated in experiments designed using the single persistent novelty setup introduced in Section \ref{sec:problem}. For each domain, we selected a few novelty instantiations that we expect the adaptation method to detect and accommodate. We ran individual experiments for each selected novelty in every domain. The experiments were set up as follows. Each trial starts with non-novel episodes, and novelty is introduced after the first $k$ episodes and persists until episode $N$ at which point the trial ends. The novelty persists until the end of the trial. After a trial ends, the agent and the environment is reset to default and the next trial commences. Each experiment was run for $L$ trials and the results below report average performance. 

\subsubsection{Accommodation through Heuristic Search-Based Repair}
First, we study the efficacy of the proposed heuristic search-based repair method for accommodation. Through empirical evidence collected from all three development domains - CartPole++, ScienceBirds, and PogoStick, we show that the proposed method:
\begin{enumerate}
    \item Maintains resilience of planning-based agents;
    \item Learns quickly with comparably fewer interactions with the environment;
    \item Is interpretable by design;
    \item Is general: can address a variety of novelties including those that negatively affect performance and those that create an opportunity to improve it; 
    \item Is domain-independent: adapts planning models for multiple domains.
\end{enumerate}

\paragraph{CartPole++} 
In CartPole ++, the novelty we study is increasing the mass of the cart by a factor of $10$ (novelty ID 1 in Table \ref{tab:novelties}). This novelty was selected because it significantly affects the dynamics of the CartPole++ system, making it less controllable without understanding the impact of the novelty. We compare the performance for four different agents: planning-static, planning-adaptive (HYDRA), DQN-static, and DQN-adaptive. We describe these agents below. For each agent, we ran $R=10$ trials with $N=30$ episodes and novelty was introduced after $k=7$ episodes. 

\begin{table}[ht]
\begin{tabularx}{\columnwidth}{XXX}
\toprule
Repairable fluents       & Nominal  & $\Delta$  \\ \midrule
length pole           & 0.500     & 0.100  \\
mass pole             & 0.100     & 0.100  \\
mass cart             & 1.000     & 1.000  \\
force magnitude       & 10.000    & 1.000  \\
gravity               & 9.810     & 1.000  \\
pole angle limit      & 0.165     & 0.010  \\
push force [L,R,F,B]  & 10.000    & 1.000  \\
pole velocity [x,y]   & 1.000     & 1.000  \\
cart velocity [x,y]   & 1.000     & 1.000  \\
                            \bottomrule
\end{tabularx}

\caption{Repair parameters for CartPole++. Shown are the repairable fluents (for vectors, each element value can be adjusted separately), nominal values, and the $\pm\Delta$ MMO values used in repair. }
\label{tab:cartpole_domain_parameters}
\vspace{-3mm}
\end{table}

The \emph{planning-static} agent selects which action to perform by using a PDDL+ domain that is consistent with the environment before novelty has been introduced. \revision{In essence, the planning-static approach is the base agent without the ability to adapt to novelty, as described in Section~\ref{sec:base-agent}.}
The \emph{planning-adaptive} (HYDRA) agent initially selects which action to perform just like the planning-static agent. However, it implements the HYDRA framework and monitors plan execution, automatically repairs the PDDL+ model when inconsistency is detected using \emph{focused model repair}, as proposed in Section \ref{sec:accomodation}. The \emph{repairable fluents} are all parameters defining CartPole dynamics, summarized in Table~\ref{tab:cartpole_domain_parameters}. To estimate the inconsistency score $C$, the planning-adaptive agent uses Euclidean distance over only the pose of the CartPole (i.e., $\tuple{cart\_x, cart\_y, theta\_x, theta\_y}$). \revision{In non-novelty case $C{=}0$, while the inconsistency threshold was set to $C_{th}{=}0.009$ based on experimentation data. The planning-static and planning-adaptive agents use the same heuristic planning approach to solve the generated PDDL+ problems, they use a time discretization $\Delta t=0.02s$, and a Greedy Best-First Search algorithm. Search is guided by a domain specific heuristic: $ h(s) = \sqrt{\theta_x^2 + \theta_y^2} * (t_{limit} - t(s))$ where $h(s)$ is the heuristic estimate of state $s \in S$, $\theta_x$ and $\theta_y$ are the angles of the pole in $x$ and $y$ planes, respectively. $t_{limit} - t(s)$ is the remaining time of the episode, i.e., the current time of the considered state $t(s)$ subtracted from the duration of the episode $t_{limit} = 0.02s * 200$ steps. In other words, the heuristic prioritizes \emph{safe} configurations with small pole angles that occur closer to the end of the episode. With this heuristic, HYDRA's PDDL+ planner normally takes under 1 second to generate a valid 200-step plan for a nominal CartPole++ planning problem given a random initial configuration.}


The DQN-static and DQN-adaptive agents are pure RL agents, employing a standard deep Q-network (DQN) implementation with experience replay memory \citep{mnih2013playing}. The Q-network is built with a dense input layer ($10 \times 512$), two hidden layers ($512 \times 512$), and a dense output layer ($512 \times 5$) and uses the Rectified Linear Unit (ReLU) activation function. The Q-network was trained to achieve perfect performance in the canonical setup. The \emph{DQN-static} agent applies the policy learned in the canonical setup in the novelty setup. This baseline was implemented to ascertain that the introduced novelty indeed impacts the performance of the agent and motivates adaptation. The \emph{DQN-adaptive} agent is initialized as the DQN-static agent, training on the non-novel episodes. But, \revision{the difference is that the DQN-adaptive agent} continues to update its weights after novelty is introduced, allowing it to potentially adapt to novelties.


\begin{figure*}
\begin{center}
\includegraphics[width=0.45\textwidth]{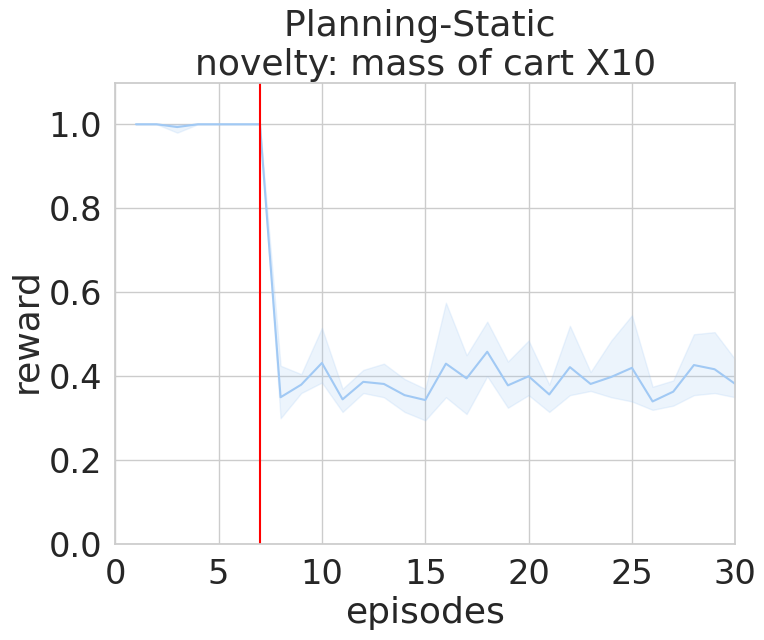}
\includegraphics[width=0.45\textwidth]{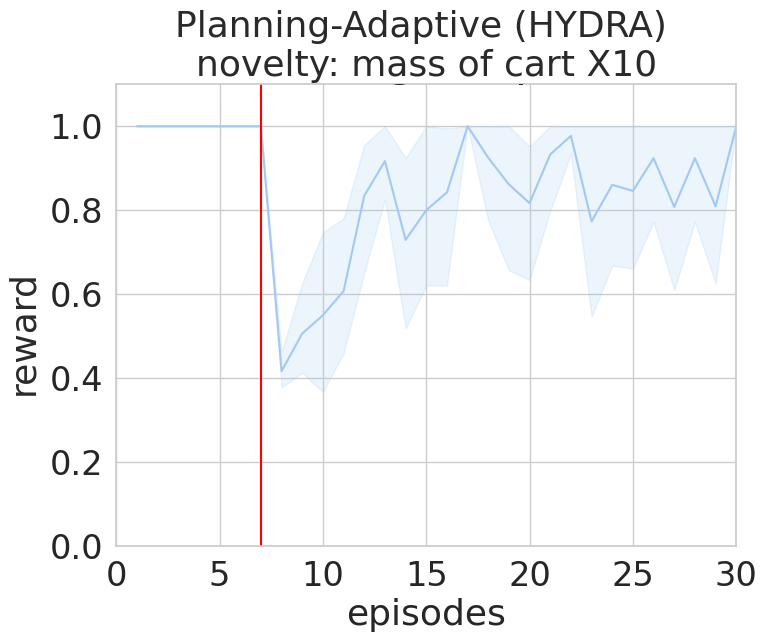}
\includegraphics[width=0.45\textwidth]{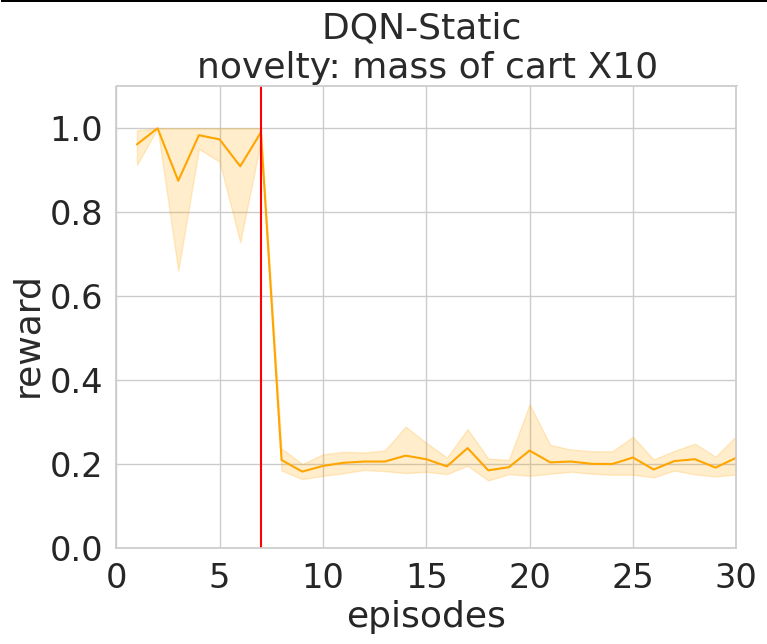}
\includegraphics[width=0.45\textwidth]{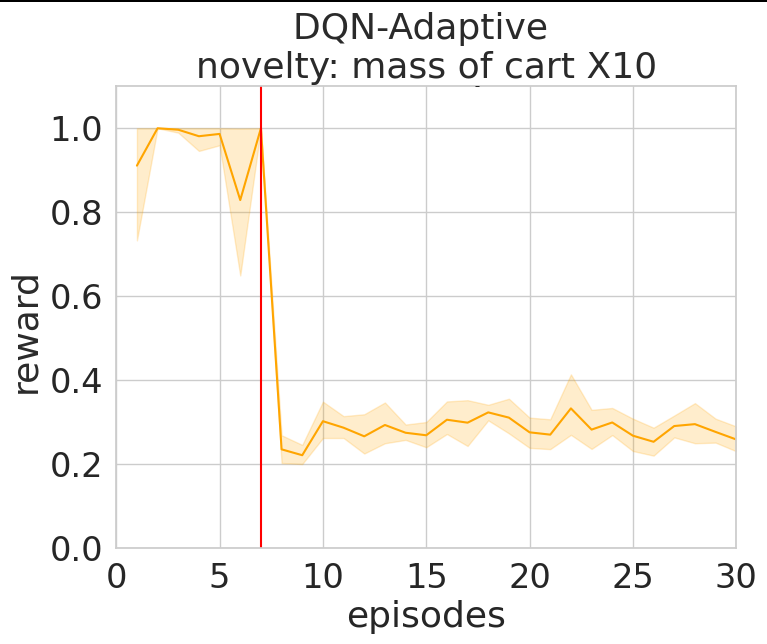}

\end{center}
\vspace{-3mm}
\caption{Adaptation performance of the static and adaptive versions of planning and DQN agents in CartPole++. \revision{The planning-adaptive approach is the HYDRA agent.} Red line denotes the introduction of novelty which \emph{increases the mass of the cart by a factor of 10}.}
\label{fig:combined_results}
\vspace{-5mm}
\end{figure*}

The performance of the planning and DQN agents is summarized in Figure \ref{fig:combined_results}. In each graph, the $x$-axis captures the episodes in a trial and the $y$-axis shows the total reward collected by the agent per episode (normalized to \tuple{0.0,1.0}). The red line indicates the episode where the novelty (shown in the graph title) was introduced. The shaded area represents the $95\%$ confidence interval computed over all 10 trials. 

As shown in Figure \ref{fig:combined_results}, all agents demonstrate perfect or near-perfect performance at the beginning of the trial and then experience a significant drop in performance when novelty is introduced (episode $8$). This drop demonstrates that the changes in the environment dynamics impact the performance of all agents. There is variability in how each agent responds to the changes in the environment. We make the following observations:

\begin{figure*}
\begin{center}
\includegraphics[width=0.32\textwidth]{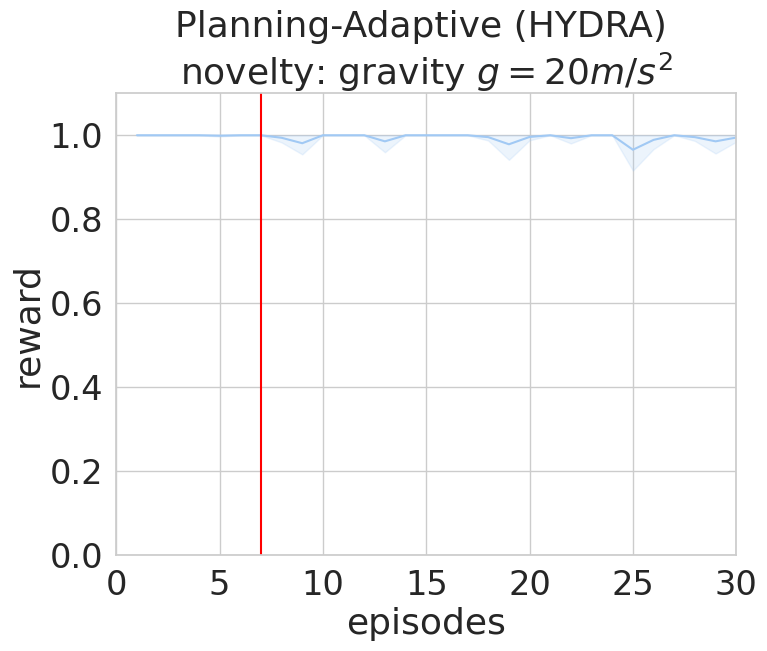}
\includegraphics[width=0.32\textwidth]{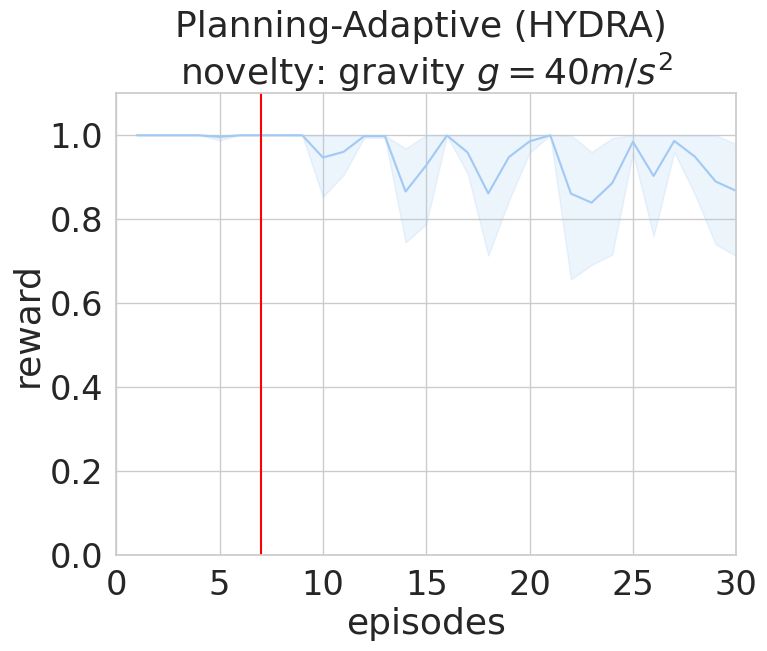}
\includegraphics[width=0.32\textwidth]{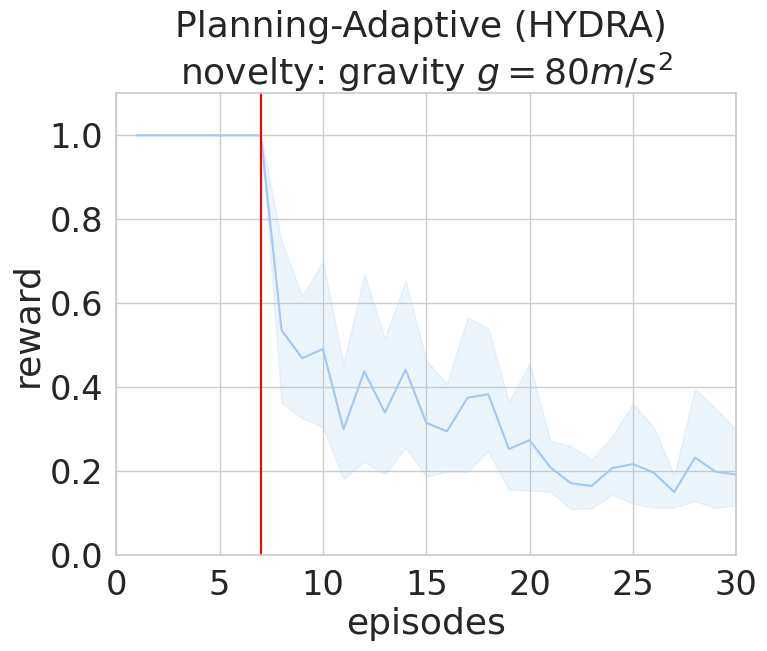}
\end{center}
\vspace{-3mm}
\caption{Adaptation performance of HYDRA agent under novelty in CartPole++. The red line denotes the introduction of novelty which adjusts the gravity to $g=20$ and $g=-40$.}
\label{fig:results_gravity}
\end{figure*}

\emph{Resilience of planning agents}: The novelty-induced performance drop is more significant in learning agents. After the introduction of novelty, the performance of the DQN agents drops to approximately $20\%$ for DQN-adaptive and DQN-static. In contrast, the performance of the planning agents only drops to $\approx40\%$. This difference can be explained by the agents' design. The planning agents' PDDL+ model defines the system dynamics in a general manner. Thus, it can still be sufficiently accurate in some conditions, even if some part of it is inaccurate. On the other hand, the DQN agent's learned policy is not general and is only applicable to a much-reduced subset of cases after novelty. 

\revision{Resilience of the PDDL+ planning agents is further exemplified by results shown in Figure~\ref{fig:results_gravity}, which showcases HYDRA's performance on novelty which alters the force of gravity (novelty ID 6 in Table \ref{tab:novelties}). The novelty increases the force of gravity from the default value of $g=9.81\frac{m}{s^2}$ to $g=20\frac{m}{s^2}$, $g=40\frac{m}{s^2}$, and $g=80\frac{m}{s^2}$, respectively. The first domain shift ($g=20\frac{m}{s^2}$) does not cause a significant drop in HYDRA's performance and the agent does not attempt any model repairs. HYDRA is also robust against the second novelty instance ($g=40\frac{m}{s^2}$), though its overall performance is only marginally worse than under the first novelty instance. HYDRA makes a single attempt to repair in 30\% of the trials, indicating that the agent's internal PDDL+ planning model remains sufficiently accurate after the domain shift. The final novelty instance ($g=80\frac{m}{s^2}$) exerts an overwhelming force on the cart and pole, which renders it uncontrollable. HYDRA persistently attempts to repair the agent's internal model but cannot find an appropriate update to make the system controllable again. This set of results shows the robustness and resilience of HYDRA, and specifically its underlying explicit planning model which can preserve accuracy even under some classes of impactful novelties.}
    
\emph{Quick adaptation via model-space search}: As expected, after novelty is introduced, the static versions of the DQN and planning agents continue performing poorly, while the adaptive agents improve their performance over time. However, the time taken to improve differs greatly between the DQN-adaptive and planning-adaptive (HYDRA) agents. Learning in DQN-adaptive is slow, requiring multiple interactions with the environment. In fact, post-novelty DQN-adaptive took $300$ episodes to increase $\approx10\%$ to reach $\approx40\%$ of optimal performance. In contrast, the planning-adaptive (HYDRA) agent recovers very quickly to $\approx100\%$ in as few as $6$ episodes. This observation supports our central thesis: model-space search enables quick adaptation in dynamic environments because it can localize the learning to specific parts of the explicit model. Other parts of the explicit model are directly \emph{transferred} to the novel setup. Knowledge transfer is challenging to implement in model-free methods (e.g., DQNs) in which action selection knowledge is distributed through the network. \revision{Furthermore, the repair module takes seconds to find an adequate model update. In the increased mass of cart example shown in Figure~\ref{fig:combined_results}, the default search-based model repair approach finds adequate model modifications to accommodate the novelty in $19s$, whereas the focused model repair only took $1.34s$ on average\footnote{Cartpole++ experiments were conducted on a Dell Precision 7560 laptop with 16-core Intel i7 processor and 64GB of RAM, running an Ubuntu 20.04 OS.}.}

\begin{figure*}
\begin{center}
\includegraphics[width=0.45\textwidth]{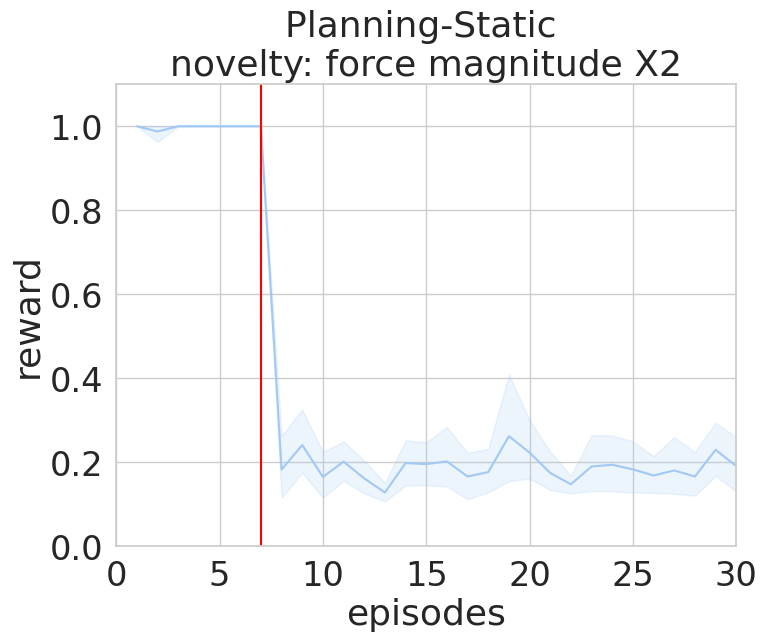}
\includegraphics[width=0.45\textwidth]{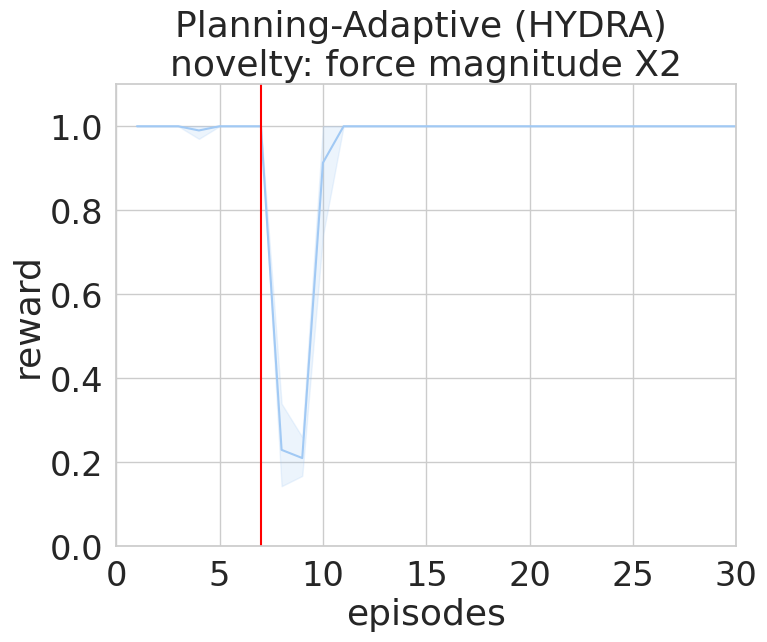}
\includegraphics[width=0.45\textwidth]{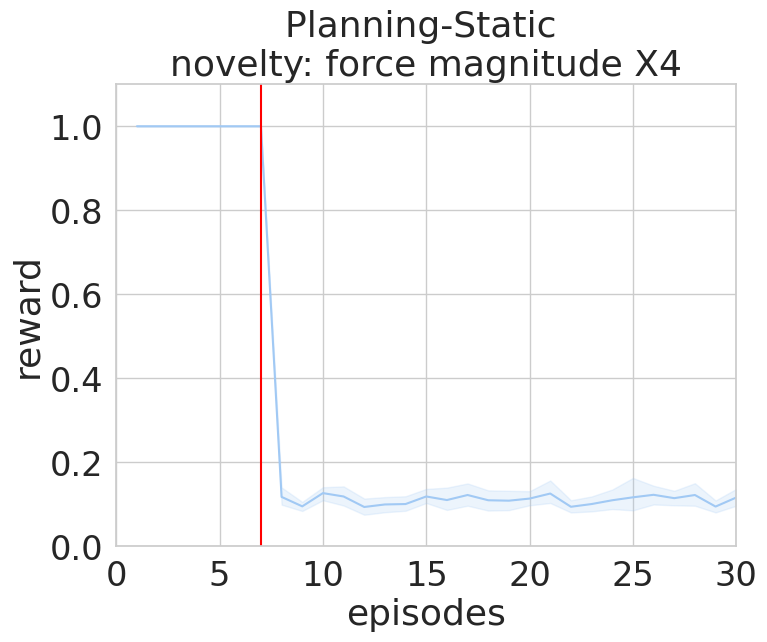}
\includegraphics[width=0.45\textwidth]{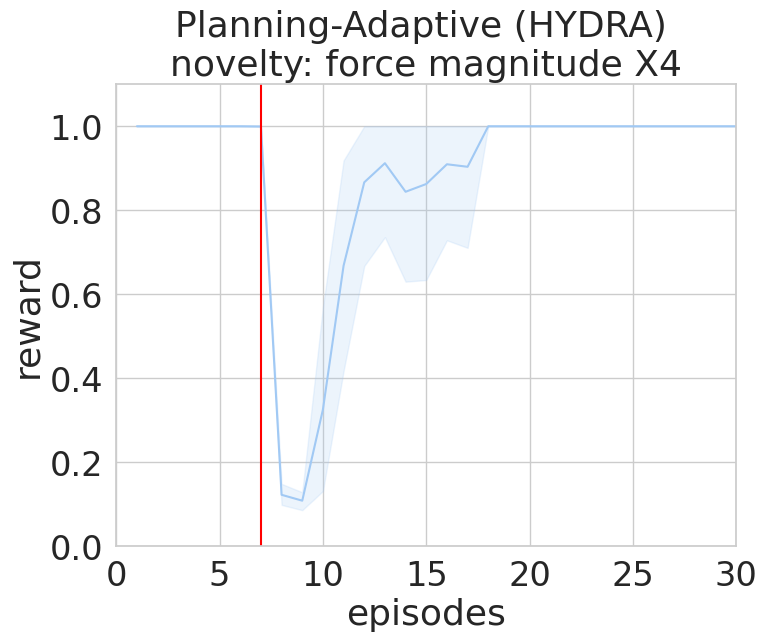}
\end{center}
\vspace{-3mm}
\caption{\revision{Adaptation performance of the planning-static base agent (left subfigures) and planning-adaptive HYDRA agent (right subfigures) under novelty in CartPole++. The red line denotes the introduction of novelty which increases the action push force by a factor of 2 (top row) and a factor of 4 (bottom row).}}
\label{fig:results_push_force}
\end{figure*}

\revision{HYDRA's rapid and accurate adaptation is further exemplified by its performance on novelty which affects the agent's push actions. The novelty (denoted ``force magnitude" in Table~\ref{tab:cartpole_domain_parameters}) increases the force $F_{mag}$ exerted on the cart from all push actions by a factor of 2 and 4 (default push force $F_{mag}=10N$)\footnote{Note that force magnitude affects the base force push, whereas each action has an individual coefficient that allows repairing the push force in a single direction only.}. Results show that HYDRA very quickly adapts to the novelty after first detection. Figure~\ref{fig:results_push_force} depicts a comparison between the planning-static base agent and planning-adaptive HYDRA agent. It is clear that the push force domain shift significantly affects the planning agent with reward dropping approximately 80-90\% immediately after novelty injection. While the planning-static never recovers from the initial drop in performance, the planning-adaptive HYDRA agent adapts very quickly and achieves perfect performance within only a couple of episodes. In the first experiment where the push force was increased two-fold, HYDRA consistently found adequate repairs (estimating the novelty-adjusted push force to be $F_{mag}{=}20N$ or $F_{mag}{=}21N$). 
The results for the 4-fold increase in the force push magnitude, show that HYDRA recovers nominal performance within a few episodes after novelty is introduced. Interestingly, repairing the agent's model using the execution trace from a single episode may sometimes cause HYDRA to find an erroneous initial repair. However, given further evidence from subsequent episodes, HYDRA can correct its erroneous model updates and converge on a correct repair. This behavior is depicted by the high spread of the shaded confidence interval (episodes 11-17) in the bottom-right planning-adaptive plot in Figure~\ref{fig:results_push_force}.}
\medskip
    
\emph{Interpretable by design}: The model repair mechanism proposes specific and localized changes to the agent's explicit PDDL+ model. Thus, adaptation in the HYDRA agent is \emph{interpretable}. A model designer can inspect the proposed repair to understand why and how the novelty affected the agent's behavior. Here is a repair found by the method during evaluation: \begin{small}\begin{spverbatim}repair:[m_cart: 9.0, l_pole: 0, m_pole: 0, force_mag: 0, gravity: 0, ...]; resulting inconsistency: 0.0067561\end{spverbatim}\end{small}.

In contrast, learning in model-free systems such as DQN-adaptive cannot be interpreted directly. The planning-adaptive (HYDRA) agent uses only its observations over a single episode in the environment to guide its search for the correct model. The observations by themselves may not provide sufficient information to determine the parameter values exactly. The model repair mechanism might be repeated after different episodes to further update the model and increase its accuracy given new trajectories. This is occasionally seen in this experimental evaluation when the PDDL+ model of the planning-adaptive (HYDRA) agent is updated multiple times, each bringing it closer to the ``true'' repair (mass of cart $\times10$).

\begin{figure*}
\begin{center}
\includegraphics[width=0.45\textwidth]{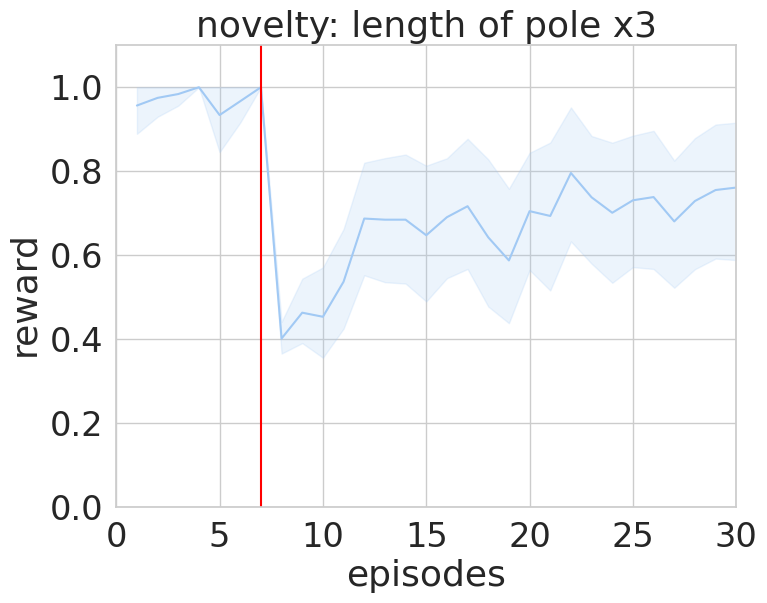}
\includegraphics[width=0.45\textwidth]{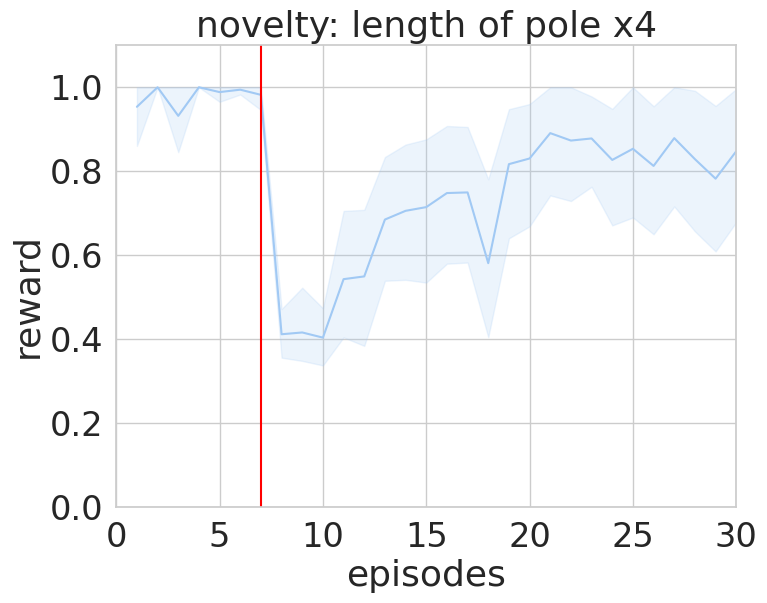}
\end{center}
\vspace{-3mm}
\caption{Adaptation performance of HYDRA agent under novelty in CartPole++. The red line denotes the introduction of novelty which increases the length of the pole by a factor of 3 (left) and a factor of 4 (right).}
\label{fig:results_length}
\vspace{-5mm}
\end{figure*}

Then next set of experiments was run to evaluate the generality of the repair-based accommodation mechanism. \revision{Figure~\ref{fig:results_length} summarizes the planning-adaptive (HYDRA) agent's behavior when novelty introduction changes the length of the pole (novelty ID 1 in Table \ref{tab:novelties})}. The results show that changing the length of the pole is impactful and degrades the agent's performance (as evidenced by the performance dip immediately after the novelty is introduced) but the repair mechanism enables the agent to recover its performance to a significant extent. 

\paragraph{ScienceBirds}

\begin{table}[ht]

\begin{tabularx}{\columnwidth}{XXX}
\toprule
Repairable fluents       & Nominal  & $\Delta$  \\ \midrule
gravity           & 9.8    & 0.100  \\
bird velocity change             & 0     & 1  \\
pig life multiplier            & 0.0     & 50  \\
                            \bottomrule
\end{tabularx}

\caption{Repair parameters for ScienceBirds. Shown are the repairable fluents (for vectors, each element value can be adjusted separately), nominal values, and the $\pm\Delta$ MMO values used in repair. }
\label{tab:domain-paramaters-sb}
\vspace{-3mm}
\end{table}

In ScienceBirds, we studied the novelty that increases the gravity in the environment (novelty ID 6), which causes the bird to fall short of its target. A planning-adaptive (HYDRA) agent was designed with repair parameters in Table \ref{tab:domain-paramaters-sb}. The inconsistency score was computed as described in Example \ref{sec:ex:inconsistency} and the threshold $C_{th}$ was set to $10$. \revision{HYDRA uses a domain-specific heuristic that considers the proximity of the in-flight bird to pigs and prioritizes states with bird trajectories that collide with pigs~\citep{piotrowski2023heuristic}. The PDDL+ planner uses a time discretization of $\Delta t = 0.025s$ and a runtime limit of 30 seconds per PDDL+ problem.} In the experiment, we ran $R=5$ trials with $N=30$ episodes and novelty was introduced after $k=7$ episodes.

\begin{figure*}
\begin{center}
\includegraphics[width=0.45\textwidth]{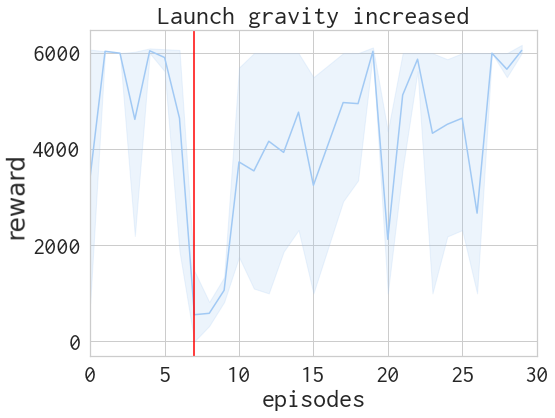}
\includegraphics[width=0.45\textwidth]{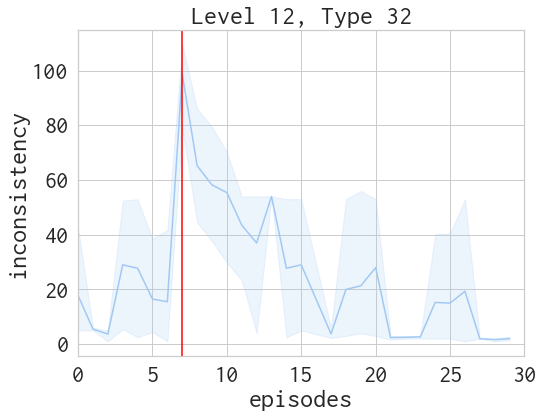}
\end{center}
\vspace{-3mm}
\caption{Adaptation performance of planning-adaptive (HYDRA) agent in ScienceBirds on novelty that increases the launch gravity of birds. The red line denotes the introduction of novelty.}
\label{fig:combined_results_sb}
\vspace{-5mm}
\end{figure*}

Results from this experiment are shown in Figure \ref{fig:combined_results_sb}, left. There are a few key observations to make. First, the performance of the agent in non-novel levels (i.e., before the novelty was introduced in episode $7$) is not perfect and it misses passing the level in some instances. ScienceBirds is a complex, continuous domain and even small inaccuracies in modeling or errors in observations can lead to failures during plan execution. After the novelty is introduced, the performance of the agent drops significantly demonstrating that the novelty was impactful. However, we see that the repair mechanism is triggered as the inconsistency score increases beyond the set threshold. The repair is able to change the gravity parameter to improve the agent's performance. There is significant variability in how soon the relevant repair is found in different trials. However, by the 28th episode, the agent has recovered its performance. Correspondingly, we see (Figure \ref{fig:combined_results_sb} right) that computed inconsistency is low in the beginning episode. As soon as novelty is introduced, we see an immediate rise in the inconsistency score which is reduced as the agent's model is autonomously repaired. This result demonstrates that the inconsistency score is sensitive to a class of novelties and can be used to detect their existence. 

\paragraph{PogoStick}
\revision{This is a complex grid-world-based domain that exhibits partial observability and contains multiple volitional agents acting in the world. 
In general, the plan to solve the episode involves exploring the environment (including unknown adjacent rooms that may contain useful items), interacting with trader agents, and gathering crafting resources, all the while avoiding being impeded by the rival pogoist agent. 
To craft a pogo-stick, the agent must first craft intermediate items, such as a tree tap for gathering rubber from a tree. The agent has knowledge about recipes to craft different intermediate items or may interact with the trader agent to obtain the recipes. The agent has an allotted budget of 2000 actions or 3 minutes per episode. }

\revision{On average, HYDRA takes approximately 200 steps and 30 seconds to solve a non-novel episode. Internally, we limit the planning time to 30 seconds per task, if no plan is found within that time, the agent starts exploring the environment to expand their knowledge base before replanning. HYDRA's planner uses a domain-specific heuristic that prioritizes states with minimized difference between the agent's inventory and the quantities of required items and resources specified by the relevant crafting recipes. }

\revision{We studied a novelty that increases the number of resources (called logs) spawned from breaking down objects in the environment. The novelty has 3 different levels -- easy, medium, and hard, based on the difficulty to detect the change in the environment and complete the required tasks. Easy novelty increases the number of logs produced 5 times the nominal quantity, medium novelty increases the number of logs 3 times, and hard novelty increases the number of logs 2 times. This novelty is an instantiation of ID $4$ (Interaction) where the agent's action of breaking down an object has changed effects, as defined in Table \ref{tab:novelties}. Table \ref{tab:domain-paramaters-polycraft} describes the MMOs and corresponding deltas that were implemented. Inconsistency threshold for PogoStick was set at $C_{th} = 2$. }

\revision{In PogoStick, the scoring system assumes a fixed cost is associated with every action of the order of magnitude of a thousand points, such as, cutting a tree to construct logs costs roughly $4000$ points. Once the agent achieves the goal, i.e. it constructs a PogoStick, the agent is rewarded $128,000$ points. If the agent gives up or does not complete the episode in time, the episode terminates with a reward of $64,000$. The final score is calculated by subtracting the action costs from the reward. 
In contrast to CartPole++ and ScienceBirds, instead of degrading performance, the introduced PogoStick novelty presents the agent an opportunity to earn a higher score by enabling them to find a shorter plan with lower total action costs.}

\begin{table}[ht]

\begin{tabularx}{\columnwidth}{XXX}
\toprule
Repairable fluents       & Nominal  & $\Delta$  \\ \midrule
break\_log           & 2    & 1  \\
break\_platinum             & 1     & 1  \\
break\_diamond            & 9     & 1  \\
collect\_saplings & 1 & 1 \\
                            \bottomrule
\end{tabularx}
\caption{Repair parameters for Pogostick. Shown are the repairable fluents (for breaking logs, platinum ore, diamond ore, and collecting saplings), nominal values, and the $\pm\Delta$ MMO values used in repair.}
\label{tab:domain-paramaters-polycraft}
\vspace{-3mm}
\end{table}

\begin{figure*}
\begin{center}
\includegraphics[width=0.66\textwidth]{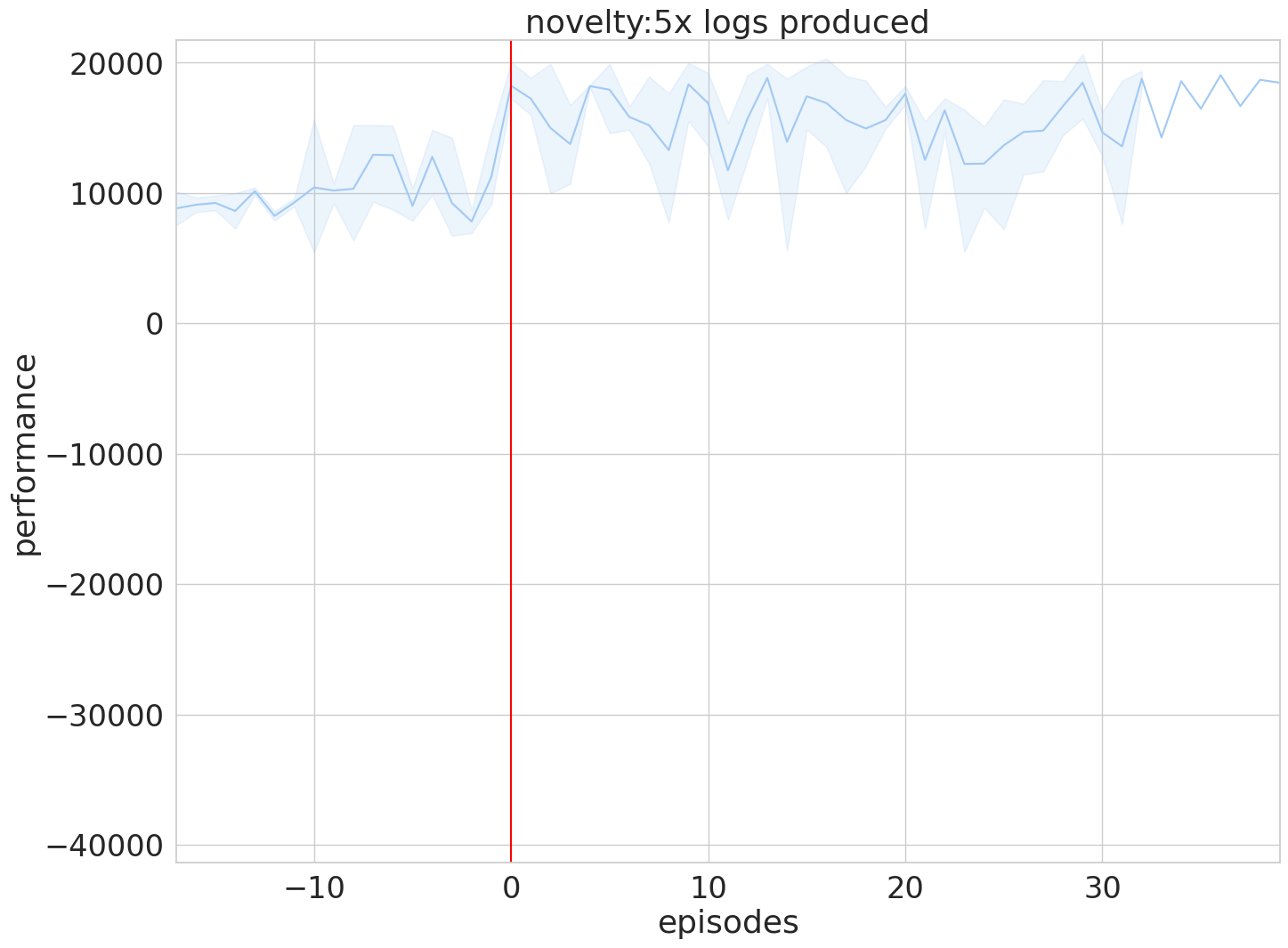}
\includegraphics[width=0.66\textwidth]{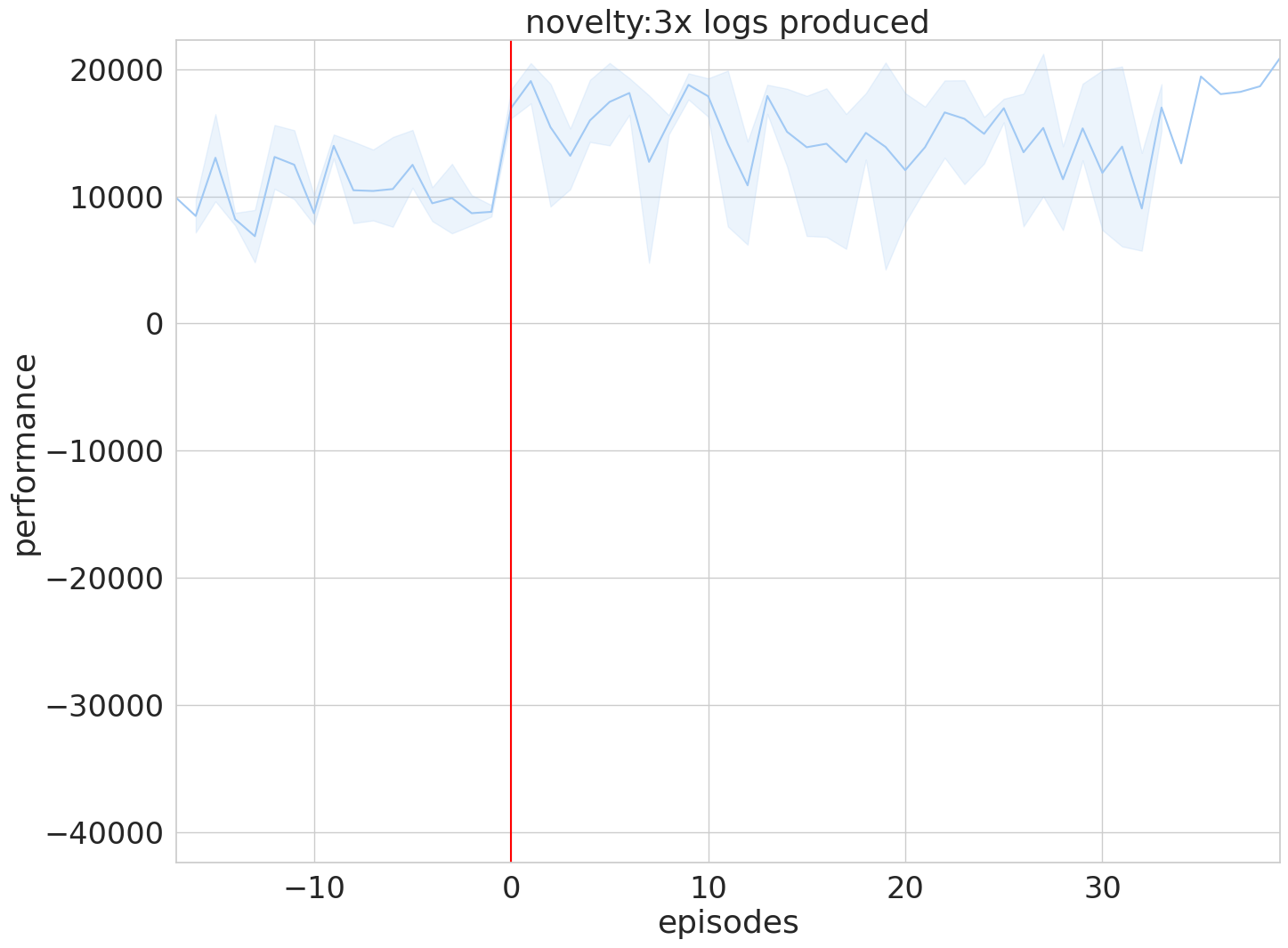}
\includegraphics[width=0.66\textwidth]{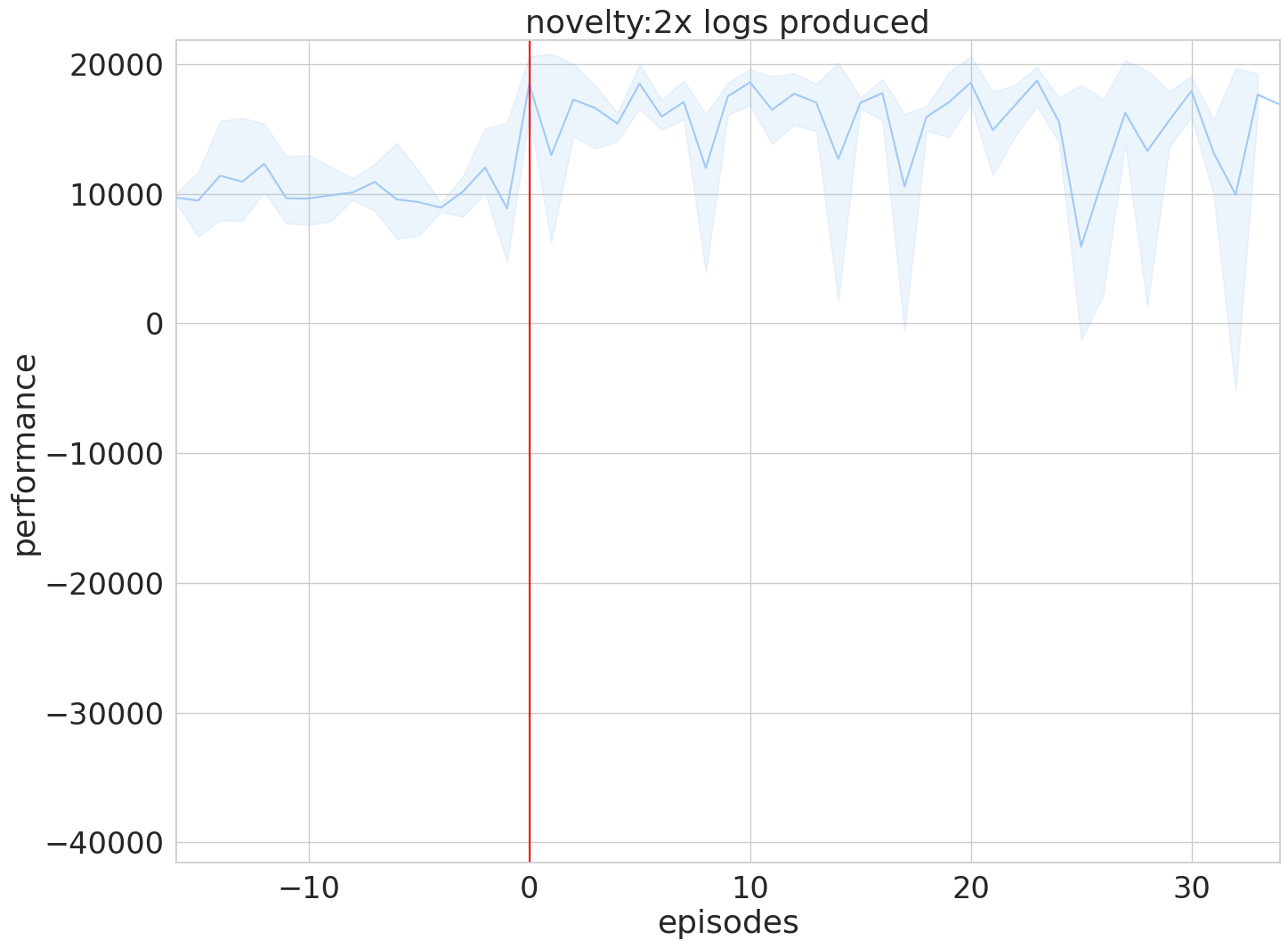}
\end{center}
\vspace{-3mm}
\caption{Result showing novelty adaptation for easy, medium, and hard novelty cases. The novelty was introduced in episode 0. The experiment consists of running three different environments provided for the specific novelty.}
\label{fig:pogostick_result}
\vspace{-5mm}
\end{figure*}

\revision{Figure \ref{fig:pogostick_result} shows the results for three different $50$ episode trials where novelty was introduced in episode $0$.
The results show that before the novelty is introduced the agent can complete the task and earn an average of $10K$ in reward. After the novelty is introduced in all three cases, the agent earns a higher reward (average of $~15K$) because now it needs fewer actions to complete the task. Soon after the introduction of novelty, in episode 1, the agent accommodates the novelty by updating its planning model and increasing the number of logs obtained. Note that the relative incorporation is instantaneous (in episode 1) as the difference between HYDRA's planning-based predictions and the execution trace is easily distinguishable and quickly updated. 
The results are consistent for all novelty difficulty levels.}

\subsubsection{Accommodation through Task Re-prioritization}
Accommodating a novelty by re-prioritizing tasks was only evaluated for ScienceBirds. CartPole++ only has one task - keeping the pole upright - and none of the novelties impact it. PogoStick does have multiple tasks but only one that pertains to the goal of the game, other tasks are exploratory tasks that enable the agent to acquire more information about its environment. These exploratory tasks don't disrupt each other. 

\begin{figure}
    \centering
    \begin{subfigure}[b]{0.45\textwidth}
    \centering
    \includegraphics[width=\textwidth]{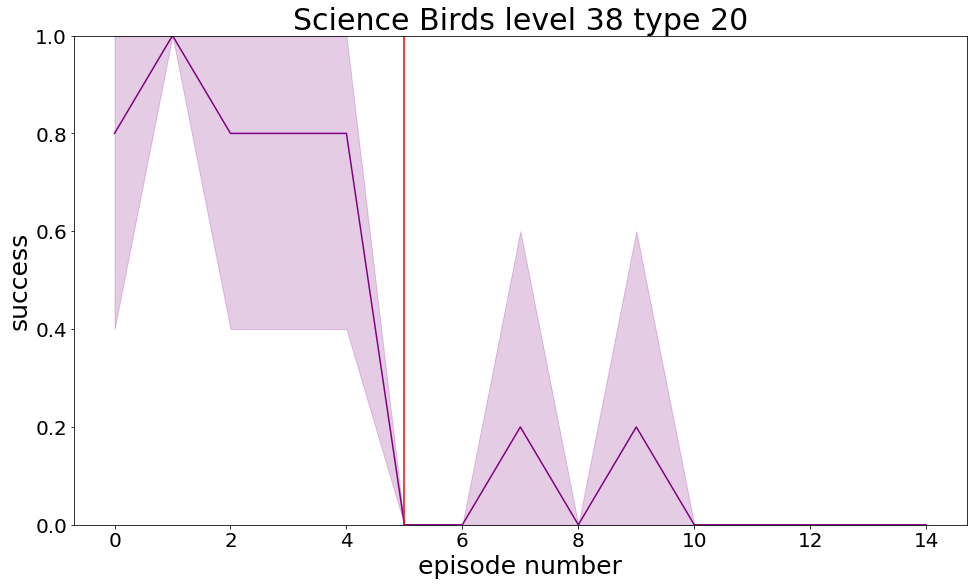}
    \caption{Without task re-prioritization}
    \label{fig:without-reorder}
    \end{subfigure}
    \begin{subfigure}[b]{0.45\textwidth}
    \centering
    \includegraphics[width=\textwidth]{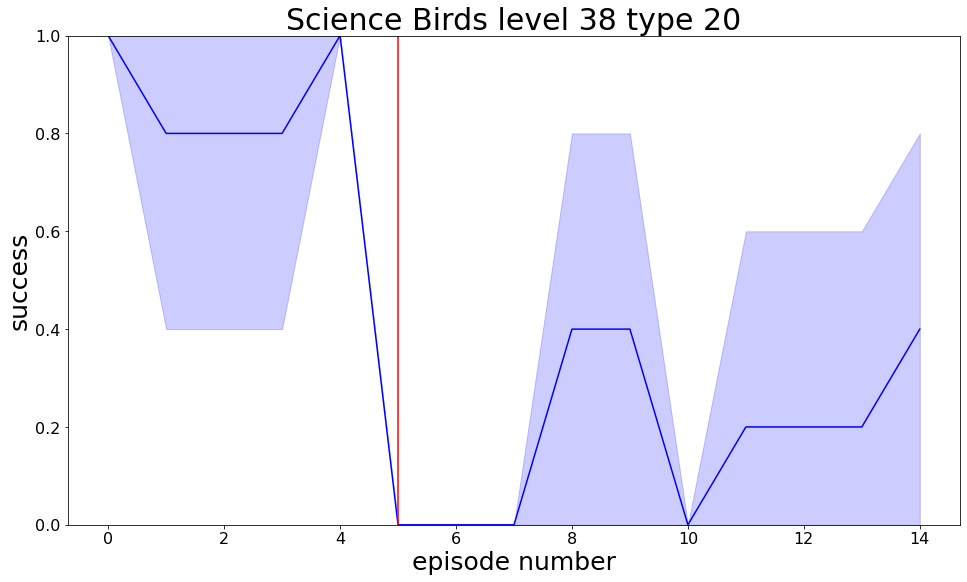}
    \caption{With task re-prioritization}
    \label{fig:with-reorder}
    \end{subfigure}
    \caption{Difference in accommodation without and with task re-reprioritization}
    \label{fig:reorder_accomodation}
\end{figure}

The effectiveness of the task re-prioritization strategy was evaluated on a specific obstructive novelty in ScienceBirds (described in Section \ref{sec:reprioritization}). In Figure \ref{fig:reorder_accomodation}, we report the performance of two various HYDRA agents: without (\ref{fig:without-reorder}) and with (\ref{fig:with-reorder}) the strategy. Along the x-axis is the episode number and along the y-axis is the observed success rate in 5 trials. 

The results show that both versions of the agent begin with similar success rates. After the novelty is introduced in episode 5, the agent without task re-prioritization degrades completely and is unable play the game successfully. The agent version with task re-prioritization is able to change the order in which it shoots at the pig and the payoff of this re-prioritization is visible as the increase in success rate towards the later parts of the trial. An interesting question here is why even after changing the task order, the agent is unable to get close to a $1.0$ success rate, similar to that in the non-novel condition. The answer lies in the uncertainty of action execution in ScienceBirds. It is more challenging to shoot at the pig on the platform and HYDRA is only able to make a successful shot $\approx 50\%$ of the time. This uncertainty is reflected in the success rate. The setup only has two birds. If one bird fails to hit the target pig, the game is lost. 

Further examination of the shots made by the agent revealed that it indeed made the shots in the right order.

\subsection{Summary of Results}

Table~\ref{tab:results_summary} provides an overview of our experimental results with respect to the types of novelties considered in this work (Table~\ref{tab:novelties}). 
Rows in Table~\ref{tab:results_summary} represent novelty types and columns represent detection or adaptation tasks (``D'' or ``A'') for each domain (Cartpole++, ScienceBirds, and PogoStick). 
In each cell, we mark whether the current implementation of HYDRA shows positive results for detecting or adapting to novelties of the corresponding tasks. The value in each state is ``$\checkmark$'', ``*'', or ``.'' representing showing positive results, not showing positive results, and settings not evaluated as of now, respectively. The resulting overview shows a promising outlook: the proposed domain-independent framework can detect and accommodate novelty IDs $1$ and $6$ and additionally detect IDs $2$ and $4$ in at least one domain. As we extend HYDRA to incorporate other model repair and learning mechanisms, we expect it to cover a larger space of novelties.


\begin{table}[ht]
\centering
\footnotesize
\begin{tabular}{@{}llp{4cm}cccccc@{}}
\toprule
ID &
  Type &
  Description &
  \multicolumn{6}{c}{Evidence} \\
 &
   &
   &
  \multicolumn{2}{c}{CartPole++} &
  \multicolumn{2}{c}{ScienceBirds} &
  \multicolumn{2}{c}{PogoStick} \\ 
&&&
\multicolumn{1}{c}{D}&
\multicolumn{1}{c}{A}&
\multicolumn{1}{c}{D}&
\multicolumn{1}{c}{A}&
\multicolumn{1}{c}{D}&
\multicolumn{1}{c}{A} \\
  \midrule
1 &
  Attribute &
  New attribute of a known object or entity & 
  \checkmark & \checkmark &
  \checkmark & \checkmark &
  \checkmark & \checkmark 
   \\
2 &
  Class &
  New type of object or entity &
  \checkmark & . &
  \checkmark & . &
   \checkmark & .\\
3 &
  Action &
  New type of agent behavior/control &
  * & * &
  * & * & 
  * & * \\
4 & Interaction  & 
  New relevant interactions of agent, objects, entities & 
  \checkmark & . & 
  \checkmark & . &  
  \checkmark & \checkmark \\
5 &
  Activity &
  Objects and entities operate under new dynamics/rules &
  \checkmark & . &
  \checkmark & \checkmark &
  . & . \\
6 &
  Constraints &
  Global changes that impact all entities &
  \checkmark & \checkmark &
  \checkmark & \checkmark &
  \checkmark & \checkmark 
   \\
7 &
  Goals &
  Purpose of the agent changes &
  * & * &
  * & * &
  * & * \\
8 &
  Processes &
  New type of state evolution not as a direct result of agent or entity action &
  \checkmark & . &
  . & . &
  . & . \\ 
  \bottomrule
\end{tabular}

\caption{Summary of HYDRA's performance on various types of novelties in our research domains. D stands for detection capability and A stands for characterization and accommodation capabilities. \checkmark denotes that HYDRA shows evidence, . denotes that HYDRA doesn't yet show evidence, and * denotes that there is no instantiation.}
\label{tab:results_summary}
\end{table}
\section{Conclusion and Future Work}
\label{sec:conclusions}
Despite employing a range of computational methods - from hand-designed model-based reasoning systems to model-free learning systems - autonomous agents depend upon the availability of accurate models of the environment during design time \citep{langley2020open}. While model-based reasoning agents (e.g., planning) encode the models explicitly, model-free learning agents (e.g., reinforcement learning) learn action selection based on these models (available via simulations). This assumption of a closed world is unlikely to hold when the agents are deployed in real-world environments. Either because the real world was not correctly understood to develop the models or because the real world evolves. Agents that can robustly handle open worlds - environments that may change while the agent is operational - have been recently proposed as a challenge for intelligent system design \citep{langley2020open, AI_for_openworlds_2022}. 

This paper introduces HYDRA - a domain-independent framework for implementing agents that are \emph{novelty-aware} - i.e. they can detect, characterize, and accommodate novelties during performance. HYDRA is built upon model-based reasoning and exploits the explicit and compositional nature of knowledge implemented in such systems. HYDRA frames learning as a volitional meta-reasoning activity that monitors the agent's own behavior, identifies an opportunity when it diverges from what is expected, and adapts the models that underlie action reasoning. The framework leverages a wide array of computational methods including continuous domain planning, classical and deep learning, heuristics search, diagnosis, and repair. HYDRA contributes to the growing instances of integrated intelligent systems that employ a variety of intelligent algorithms in a single, end-to-end architecture that demonstrates complex behavior. 

We used HYDRA to design novelty-aware agents for three complex, mixed discrete-continuous domains: CartPole++, ScienceBirds++, and PogoStick. We found that the method can be implemented in a domain-independent fashion with few domain-specific elements that guide search and establish a success criterion. Through empirical analyses, we show that adaptation by model repair and task re-prioritization enables the agents to accommodate novelties. Model repair can occur rapidly, requiring only a few interactions with the environment. Further, revisions are expressed in the language the model is written in and therefore, are interpretable by design. A model designer can inspect what the agent learns. Our analyses show that the proposed method retains the strengths of model-based reasoning methods (structure and explicability) while making them adaptable to changing environmental dynamics. We also demonstrate that learning by model repair can alleviate some challenges inherent in modeling a complex domain - if the domain designer writes an inaccurate model, the proposed method can iterate it based on observations from the environment. 

Recently, \cite{chao2023novelty} designed and evaluated a HYDRA agent on a high-fidelity wargaming simulator highlighting that the HYDRA framework is domain-independent and supports the design of open-world learning agents for a variety of environments. 

\subsection{Limitations}
\revision{While HYDRA takes some significant strides towards designing agents that can operate in open worlds, several gaps must be addressed before such agents can be deployed in the real world. The presented environments push on some dimensions of environmental complexity: all of them are open, CartPole++ and ScienceBirds have mixed discrete-continuous state-action spaces but are fully observable, on the other hand, PogoStick is discrete but partially observable. Other complexity dimensions such as sensing noise, non-deterministic actions, probabilistic state transitions, etc.. pose significant challenges for agent development as well as for reasoning about novelty but are not studied here.}

\revision{Further analyzing HYDRA's efficacy in handling open worlds, it is clear that the implemented methods address a subset of novelty types; a complete open-world learning agent can reason about and accommodate the entire space of novelties that are plausible in its domains. Despite this limitation, HYDRA is a framework for open-world goal-oriented agents that can be incrementally extended to include other accommodation methods. Another limitation is the assumption of a single persistent novelty setup which assumes that only one type of novelty presents and remains until the end. A more general framing of this problem would allow for multiple novelties to appear either sequentially or together, motivating continual accommodation for open worlds. We expect that HYDRA's design will facilitate more advanced forms of open-world learning.}

\revision{In its current state, the model repair method proposed in this paper has two main limitations: (1) it relies on manually selected subsets of repairable variables, as well as domain-specific repair parameters such as repair deltas or the inconsistency threshold and (2) it is restricted to model updates defined by the repairable fluents. While effective in the cases studied in this paper, this method can be further enhanced to repair other aspects of a planning model.}
\smallskip

\subsection{Future Work}
\revision{There are several avenues to directly address the aforementioned limitations in future work. We are currently developing mechanisms to improve the efficiency of search-based model repair. 
developing an extension to HYDRA that exploits the information gathered during inconsistency estimation to efficiently search the space of MMOs. As a result, HYDRA will automatically select a relevant subset of variables that can be efficiently repaired. However, for best repair performance, such developments must be paired with accurate domain-independent inconsistency checkers that will allow HYDRA to accurately compare traces regardless of the PDDL+ model and better detect novelty. We are also developing an optimization-based approach to finding the MMO $\Delta$ adjustment quantum (currently set manually a priori). Overall, these improvements will transform HYDRA into a much more domain-independent framework.}

\revision{The next avenue is expanding the classes of novelties that HYDRA can reason about and accommodate. We are exploring how the repair framework can be extended to include modifications to the structure of the PDDL domain by adding, removing, and modifying preconditions and effects, and finally adding and removing entire happenings. This line of research will bring insights from model learning research into a larger, integrated theory of planning model repair.} 

\revisiontwo{The HYDRA framework and supporting algorithms presented in this work do not presume correctness of the repaired domain model, completeness of the repair process, or any guarantees on runtime and space complexity. Such guarantees are not possible in the most general form of our problem as novelties can have arbitrary effects. However, for specific types of novelties one may be able to devise appropriate MMOs and repair algorithms that may provide such guarantees. This is another direction for future work.}

\revision{Yet another direction is automatically updating the recognition models to incorporate new entities and learn corresponding PDDL domain models from experience. Finally, we are interested in advancing this research to a continuous open-world learning setting where new novelties are introduced on the fly and the system adapts to these domain shifts during execution.}

\revision{Open world learning is an exciting new paradigm for autonomous agents. Classical machine learning makes a distinction between the training and testing phases where the agent's knowledge is updated during training time and produces behavior at testing time. Open-world learning removes that distinction and poses the challenge of autonomously detecting when there is an opportunity to learn and determining what should be learned. Our research takes some initial but important steps towards agents that can autonomously adapt to an open world, paving the way for advances on this challenging research question.}



\section*{Acknowledgements}
The work presented in this paper was supported in part by the DARPA SAIL-ON program under award number HR001120C0040. The views, opinions, and/or findings expressed are those of the authors and should not be interpreted as representing the official views or policies of the Department of Defense or the U.S. Government. The authors thank Matt Klenk for his contributions to problem and approach formulation. 




\bibliographystyle{elsarticle-harv} 
\bibliography{bibliography}

\end{document}